\documentclass[sigconf]{acmart}

\usepackage{colortbl}
\usepackage{booktabs} % For formal tables
\usepackage{algorithm}
\usepackage{booktabs}
\usepackage{amsmath}
\usepackage{graphicx}
\usepackage{subfigure}
\usepackage{amsfonts,bm}
\usepackage{multirow}
\usepackage{color}
\usepackage{braket}
\usepackage{caption}
\usepackage{mathtools}
\usepackage{enumerate}
\usepackage{enumitem}
\usepackage{empheq}
\usepackage{calc}
\usepackage{dsfont}
\usepackage{algpseudocode}
\usepackage{subcaption}
\usepackage{enumitem}
\usepackage{subcaption}
\usepackage{makecell}

\AtBeginDocument{%
  \providecommand\BibTeX{{%
    \normalfont B\kern-0.5em{\scshape i\kern-0.25em b}\kern-0.8em\TeX}}}

\copyrightyear{2026}
\acmYear{2026}
\acmConference[KDD '26]{Proceedings of the 32nd ACM SIGKDD Conference on Knowledge Discovery and Data Mining V.1}{August 09--13, 2026}{Jeju Island, Republic of Korea}
\acmBooktitle{Proceedings of the 32nd ACM SIGKDD Conference on Knowledge Discovery and Data Mining V.1 (KDD '26), August 09--13, 2026, Jeju Island, Republic of Korea}
\acmPrice{}
\acmDOI{10.1145/3770854.3785682}
\acmISBN{979-8-4007-2258-5/2026/08}

\begin{document}

\title[X-MethaneWet: A Cross-scale Global Wetland Methane Emission Benchmark Dataset]{X-MethaneWet: A Cross-scale Global Wetland Methane Emission Benchmark Dataset for Advancing Science Discovery with AI}

\author{Yiming Sun}
\affiliation{%
  \institution{University of Pittsburgh}
  \city{Pittsburgh}
  \state{PA}
  \country{USA}
}
\email{yimingsun@pitt.edu}

\author{Shuo Chen}
\affiliation{%
  \institution{Purdue University}
  \city{West Lafayette}
  \state{IN}
  \country{USA}
}
\email{chen4371@purdue.edu}

\author{Shengyu Chen}
\affiliation{%
  \institution{University of Pittsburgh}
  \city{Pittsburgh}
  \state{PA}
  \country{USA}
}
\email{SHC160@pitt.edu}

\author{Chonghao Qiu}
\affiliation{%
  \institution{University of Pittsburgh}
  \city{Pittsburgh}
  \state{PA}
  \country{USA}
}
\email{CHQ29@pitt.edu}

\author{Licheng Liu}
\affiliation{%
  \institution{University of Minnesota}
  \city{Minneapolis}
  \state{MN}
  \country{USA}
}
\email{lichengl@umn.edu}

\author{Youmi Oh}
\affiliation{%
  \institution{NOAA Global Monitoring Laboratory}
  \city{Boulder}
  \state{CO}
  \country{USA}
}
\email{youmi.oh@noaa.gov}

\author{Sparkle L. Malone}
\affiliation{%
  \institution{Yale University}
  \city{New Haven}
  \state{CT}
  \country{USA}
}
\email{sparkle.malone@yale.edu}

\author{Gavin McNicol}
\affiliation{%
  \institution{University of Illinois Chicago}
  \city{Chicago}
  \state{IL}
  \country{USA}
}
\email{gmcnicol@uic.edu}

\author{Qianlai Zhuang}
\affiliation{%
  \institution{Purdue University}
  \city{West Lafayette}
  \state{IN}
  \country{USA}
}
\email{qzhuang@purdue.edu}

\author{Chris Smith}
\affiliation{%
  \institution{NOAA Global Monitoring Laboratory}
  \city{Boulder}
  \state{CO}
  \country{USA}
}
\email{chris.c.smith@noaa.gov}

\author{Yiqun Xie}
\affiliation{%
  \institution{University of Maryland}
  \city{College Park}
  \state{MD}
  \country{USA}
}
\email{xie@umd.edu}

\author{Xiaowei Jia}
\affiliation{%
  \institution{University of Pittsburgh}
  \city{Pittsburgh}
  \state{PA}
  \country{USA}
}
\email{xiaowei@pitt.edu}

% \renewcommand{\shortauthors}{Sun et al.}
\renewcommand{\shortauthors}{Yiming Sun et al.}

\begin{abstract}
Methane (CH$_4$) is the second most powerful greenhouse gas after carbon dioxide and plays a crucial role in climate change due to its high global warming potential. Accurately modeling CH$_4$ fluxes across the globe and at fine temporal scales is essential for understanding its spatial and temporal variability and developing effective mitigation strategies. In this work, we introduce the first-of-its-kind cross-scale global wetland methane benchmark dataset (X-MethaneWet), which synthesizes physics-based model simulation data from TEM-MDM  and the real-world observation data from FLUXNET-CH$_4$. This dataset can offer opportunities for improving global wetland CH$_4$ modeling and science discovery with new AI algorithms. To set up AI model baselines for methane flux prediction, we evaluate the performance of various sequential deep learning models on X-MethaneWet. Furthermore, we explore four different transfer learning techniques to leverage simulated data from TEM-MDM to improve the generalization of deep learning models on real-world FLUXNET-CH$_4$ observations. Our extensive experiments demonstrate the effectiveness of these approaches, highlighting their potential for advancing methane emission modeling and identifying new opportunities for developing more accurate and scalable AI-driven climate models.
\end{abstract}


% %%
% %% Keywords. The author(s) should pick words that accurately describe
% %% the work being presented. Separate the keywords with commas.
% %% A "teaser" image appears between the author and affiliation
% %% information and the body of the document, and typically spans the


\maketitle

\section{Introduction}
Methane (CH\textsubscript{4}) is the second most important greenhouse gas after carbon dioxide~\cite{stocker_climate_2013}. CH\textsubscript{4} has significant implications for climate change due to its high global warming potential~\cite{bergamaschi_atmospheric_2013,neubauer2015moving}. 
Wetlands, as the largest natural source of CH\textsubscript{4}, %with the emission of 
%Wetlands 
emitted 
159 [119-203] Tg CH\textsubscript{4} yr\textsuperscript{-1} globally during 2010-2019~\cite{saunois_global_2024} and accounted for nearly one-third of total global CH\textsubscript{4} emissions\cite{jackson_human_2024}. 
Accurately modeling wetland CH\textsubscript{4} fluxes on a fine temporal scale is critical for understanding its spatial and temporal dynamics and 
informing mitigation strategies. 
Due to its relatively short atmospheric lifetime  and greater per-molecule radiative forcing 
compared to other greenhouse gases, reduction of methane is often considered an effective way of mitigating climate change. 
However, the modeling of methane emission is very challenging as 
methane emissions are highly variable in space and time, influenced by hydrological conditions, temperature, and microbial activities.

Over the past decades, physics-based (PB) models have traditionally been used for the estimate of wetland CH\textsubscript{4} emissions. PB models use mathematical equations to explain the connection between the CH\textsubscript{4} production, oxidation, and transport processes and environmental variables~\cite{tian_global_2015,zhang_emerging_2017}. For example, the Terrestrial Ecosystem Model with Methane Dynamics Module (TEM-MDM) simulates methanogenesis in the saturated zones of soil and methanotrophy in the unsaturated zones, with both processes influenced by soil temperature, moisture, and organic carbon availability~\cite{zhuang_methane_2004}. TEM-MDM also simulates methane transport through soil layers by diffusion, plant-aided transport, and ebullition, allowing for a dynamic representation of how methane moves from the soil to the atmosphere, particularly in wetland areas~\cite{zhuang_methane_2004}. By incorporating mechanistic relationships and physical laws, PB models can predict system responses to various scenarios, including those outside the range of historical observations\cite{liu_knowledge-guided_2024}. However, PB models require extensive parametrization and complex calculations, which can limit their applicability in real-time or large-scale scenarios.

% huge interest in exploring ML algorithms
More recently, there is a growing interest in using ML methods for estimating wetland CH\textsubscript{4} emission~\cite{hatala_gross_2012,irvin_gap-filling_2021,kim_gap-filling_2020,yuan_causality_2022,chen_quantifying_2024,mcnicol_upscaling_2023,peltola2019monthly,knox_fluxnet-ch4_2019}. Compared to traditional PB models, ML models are much more computationally efficient during inference while also effectively capturing complex nonlinear patterns in CH\textsubscript{4} flux data. 
% Traditional models suffer from .... Evaluation of ML models need to be focused on generalizability
However, due to the absence of large training datasets, traditional ML approaches may lack generalizability to out-of-sample prediction scenarios, e.g., testing in different regions or performing future time projections. This is further exacerbated by the spatial and temporal variability in wetland methane observations.  %It can be challenging for them to generalize in data-limited regions or future projections due to the variability and scarcity of data in wetland methane observations.
Hence, a comprehensive benchmark dataset  is urgently needed for the development of robust ML-based models.

% existing benchmark data (working)
One ML-based research direction involves employing ML models to upscale eddy covariance (EC) measurements into global wall-to-wall flux maps~\cite{yuan_causality_2022,chen_quantifying_2024,mcnicol_upscaling_2023,peltola2019monthly}. These ML models rely on globally available predictors such as meteorological data and soil properties to build empirical models at the site level before estimating CH\textsubscript{4} fluxes across all wetland grid cells. The upscaling products could serve as benchmark datasets but also come with inherent uncertainties, such as the choice of ML models, predictor variables, and forcing data. Despite their potential as benchmark datasets, the CH\textsubscript{4} modeling community still lacks a systematic benchmarking framework that offers standardized models, evaluation metrics, and datasets for rigorously comparing and assessing future work.

In this work, we introduce X-MethaneWet, the first global wetland methane dataset that integrates physical simulations generated by TEM-MDM and true observations at a daily scale. The TEM-MDM-based simulated dataset contains the simulation of CH\textsubscript{4} fluxes across  62,470 global sites, providing a comprehensive dataset for evaluating model performance under controlled conditions. The FLUXNET-CH\textsubscript{4} dataset, on the other hand, consists of observed CH\textsubscript{4} fluxes from 30 wetland sites worldwide, representing real-world conditions with varying environmental factors. 
Such an integrative dataset enables the evaluation of ML models in both emulating PB models and predicting actual methane fluxes over large spatial scales and long time periods. More importantly, we establish a standardized evaluation framework to assess the generalizability of methane prediction model across space and time. Such an evaluation framework defines specific spatial and temporal generalization test cases, along with the evaluation metrics that best fit the task of methane prediction.

Based on the established evaluation framework, we conduct comprehensive experiments using a range of ML models, which include LSTM-based models (e.g.,  LSTM~\cite{hochreiter1997long} and EA-LSTM~\cite{li2019ea}), the temporal CNN-based model~\cite{lea2017temporal}, and  Transformer-based models (e.g., Transformer~\cite{vaswani2017attention}, iTransformer~\cite{liu2023itransformer}, and Pyraformer~\cite{liu2022pyraformer}). These tests demonstrate the promise of existing ML models in capturing complex CH\textsubscript{4} emission dynamics  at the fine temporal frequency (daily scale) from  input drivers 
and predicting CH\textsubscript{4} emissions for unobserved regions and future time periods. 
The comparison among these models also allows for analyzing the gap and opportunities of current ML models in the context of methane prediction.   %capture CH\textsubscript{4} flux dynamics under both simulated and real-world settings.

In addition to the standard ML benchmark testing, our integrative dataset also enables the exploration of knowledge transfer approaches from TEM-MDM-based simulated data to facilitate capturing observed CH\textsubscript{4} flux dynamics.  This is inspired by recent advances in knowledge-guided machine learning (KGML)~\cite{karpatne_knowledge-guided_2024, willard_integrating_2022,karpatne_knowledge_2022}, which aims to leverage the complementary strengths of both PB and ML models. 
Previous work in this field has shown that pre-training an ML model with simulated data generated by PB models can facilitate learning general physical patterns. Then this pre-trained model needs only a few observations to correct the simulation bias during fine-tuning to achieve a quality model~\cite{read2019process,rasp2021data,jia2021physics}. %kraft_towards_2022,he2022improving}. 
In addition to  this pretrain-finetune approach,  
we have also applied multiple transfer learning algorithms to all  previously considered ML models 
to test their performance 
with knowledge adapted from physical simulations. 
This experiment demonstrates the effectiveness of knowledge transfer in improving the generalization capabilities of ML models, especially in data-sparse scenarios. The results highlight the potential of integrating synthetic data with real-world observations to develop more accurate, robust, and scalable AI-driven models for wetland CH\textsubscript{4} flux estimation, contributing to better climate modeling and mitigation strategies.

% The results show that \textcolor{red}{XXX}. 

Our dataset and implementation of benchmark testing have been released\footnote{\url{https://huggingface.co/datasets/ymsun99/X-MethaneWet} (Dataset) and \url{https://github.com/ymsun99/X-MethaneWet} (Code)}.
% Our dataset and implementation of benchmark testing have been released. The dataset is available at \footnote{\url{https://huggingface.co/datasets/ymsun99/X-MethaneWet}} and the code can be found at \footnote{\url{https://github.com/ymsun99/X-MethaneWet}}.
Our contributions have been summarized as follows:
\begin{itemize} [leftmargin=15pt]
% \vspace{.05in}
\item We create the first integrative global wetland CH\textsubscript{4} dataset that combines both physical simulations and true observations. Additionally, it is the first CH\textsubscript{4} dataset to provide global-scale methane simulations at daily resolution. 
% \vspace{.05in}
\item We introduce a standardized evaluation framework for testing ML models in global high-frequent methane flux prediction. This is particularly designed with test cases and evaluation metrics to validate model generalizability over space and time.
% \vspace{.05in}
\item We conduct comprehensive tests of diverse ML models and analyze their performance in both emulating the physics-based TEM-MDM model and predicting truly observed methane emissions. We further apply multiple transfer learning approaches on these ML models, which shows the opportunities in leveraging knowledge from TEM-MDM-generated physical simulations to facilitate the prediction in data-sparse scenarios. 
\end{itemize}

\section{Related Work}

\begin{figure*}[htbp]  
    \vspace{-.18in}
    \centering
    \includegraphics[width=\textwidth]{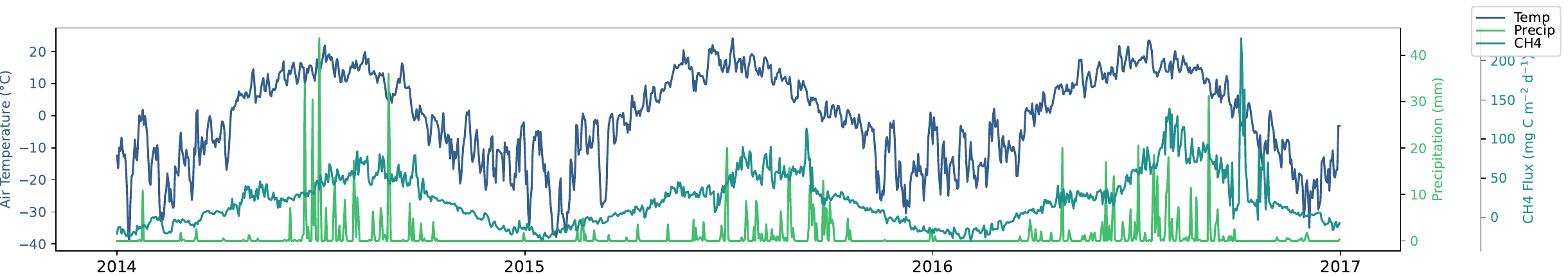} 
    \vspace{-.15in}
    \caption{Temporal variations in air temperature, precipitation, and $\text{CH}_4$ flux across 2014 to 2016 in US.BZB site. }
    \vspace{-.1in}
    \label{FLUXNET_time_series} 
\end{figure*}

% In this section, we discuss related work on  CH\textsubscript{4} benchmark datasets and the application of ML approaches to methane prediction. 

\subsection{CH\textsubscript{4} benchmark data}
Recently, significant efforts have been dedicated to developing benchmark data for CH\textsubscript{4} emissions. Most of these works utilize simple models to upscale CH\textsubscript{4} measurements to larger regions and higher resolution.  
A random forest (RF) model has been utilized to upscale CH\textsubscript{4} eddy covariance flux measurements from 25 sites for the northern region~\cite{peltola2019monthly}. Also, a causality-guided long short-term memory (causal-LSTM) model was developed and utilized data from 30 eddy covariance towers across four wetland types: bog, fen, marsh, and wet tundra~\cite{yuan_causality_2022}. 
The causal-LSTM outperformed the traditional LSTM by integrating causal relationships derived from transfer entropy analysis into the modeling process. The upscaled product is at 0.5$^\circ$ spatial resolution and covers the northern boreal and arctic regions. The time period is from 2000 to 2018 at weekly temporal resolution. The aforementioned datasets primarily focus on the Northern boreal and arctic regions. The first global benchmark dataset (UpCH\textsubscript{4}) was produced using an RF model that trained from 45 FLUXNET-CH\textsubscript{4} sites~\cite{mcnicol_upscaling_2023}. UpCH\textsubscript{4} is at 0.25° spatial resolution and global spatial coverage. The time period is from 2000 to 2018 at monthly temporal resolution. Later, another global CH\textsubscript{4} benchmark data was produced by multi-model ensemble mean of six ML models, including decision tree, RF, XGB, ANN, Gated Recurrent Units, and LSTM~\cite{chen_quantifying_2024}. This dataset was upscaled from 82 eddy covariance and chamber sites. It is at 0.5° spatial resolution and global spatial coverage. The time period is from 2000 to 2020 at monthly temporal resolution. Though great efforts were made to develop global methane benchmark datasets, no existing work has produced datasets at a daily resolution, nor clearly outlined the benchmarking  protocols, prohibiting the broad implementation of those datasets, especially for AI model development. 
Different from these datasets, we provide the first integrative dataset that consists of both simulated data and true observation data. The simulations are at high spatial resolution, and true observation data are from multiple sites worldwide at the daily scale.  Meanwhile, we provide a standardized evaluation framework to accelerate future AI model development for estimating methane emission.

\subsection{Machine learning for CH\textsubscript{4}  prediction}
Multiple ML models for estimating CH\textsubscript{4} emissions have been developed and tested in existing research. The artificial neural networks (ANNs) was utilized in the early research to predict CH\textsubscript{4} emissions in drained and flooded agricultural peatlands in the Sacramento-San Joaquin Delta~\cite{hatala_gross_2012}. The ANN model was also utilized to predict CH\textsubscript{4} emissions across 60 FLUXNET-CH\textsubscript{4} sites ~\cite{knox_fluxnet-ch4_2019}. Furthermore, the performance of multiple ML models, including ANNs, random forests (RF), and support vector machines (SVM)~\cite{kim_gap-filling_2020} were compared in the later research. RF model was found to outperform the other models across various test scenarios and sites. Another research compared the performance of Lasso regression (Lasso), ANNs, RF, and extreme gradient boosting (XGBoost). RF also outperformed the other ML models in terms of normalized mean absolute error (nMAE) and coefficient of determination (R\textsuperscript{2}) across 17 FLUXNET-CH\textsubscript{4} sites~\cite{irvin_gap-filling_2021}. However, the ML models tested in the previous research primarily focus on gap-filling rather than benchmarking.

%\subsection{Deep learning models for time-series forecasting}

% other potential ML mehtods

%In recent years, deep learning models have found significant success in time-series forecasting, particularly in capturing complex temporal dependencies and improving predictive accuracy. 
%More recently, 
%Many of these models 
%researchers have  explored other deep learning models %in time series prediction, and have shown encouraging results 
Many other deep learning models have shown encouraging results in ecological modeling %and have shown encouraging results. Hence, these models  %and thus 
and thus can also be considered promising candidates for methane emission prediction. 
For example, 
long short-term memory and its variants, such as EA-LSTM~\cite{kratzert2019towards}, incorporate gating mechanisms to capture long-term dependencies in sequential data. They are reported to outperform many advanced models in simulating ecosystem dynamics ~\cite{liu2024probing,yu2024adaptive}. %Despite their superior performance, they tend to be computationally inefficient due to their autoregressive nature in inference. 
Temporal convolution networks (TCN)~\cite{lea2017temporal} leverage causal and dilated convolutions to model long-range dependencies,  % and can be computed in parallel over the sequence. 
%Such model 
and have  also been used in many environmental applications~\cite{topp2023stream,sun2021explore,pelletier2019temporal}. % while enabling parallel processing, making it more efficient for time-series forecasting. 
Transformer-based models, such as the vanilla Transformer~\cite{vaswani2017attention,gao2022earthformer}, iTransformer~\cite{liu2023itransformer}, and Pyraformer~\cite{liu2022pyraformer}, use self-attention mechanism and have gained attention in many long-term forecasting tasks such as 
predicting weather and climate changes, as well as energy potential. %These models are able to % due to their ability to 
%capture long-range dependencies using self-attention mechanisms. 
%, and    are highly scalable as they allow 
%all time steps to be processed in parallel. %, significantly improving scalability. 
%Additionally, specialized Transformer architectures have been developed for time-series forecasting, with sparse attention mechanisms and hierarchical structures to enhance efficiency and interpretability. 
Despite the promise of advanced ML models, they often lack generalizability to data-sparse scenarios. Recent advances in knowledge-guided machine learning (KGML)~\cite{karpatne_knowledge-guided_2024, willard_integrating_2022,karpatne_knowledge_2022} address this issue by leveraging accumulated scientific knowledge to guide the learning process of ML models. Prior work in KGML has shown that the 
pretraining ML models on PB simulated data can help significantly enhance the predictive performance in monitoring poorly-observed ecosystems~\cite{read2019process,rasp2021data,jia2021physics}. 
However, similar approaches have yet been developed for estimating wetland CH\textsubscript{4} emissions.

% provide a powerful framework for integrating PB models with ML models, combining the strengths of both approaches. PB models incorporate well-established physical principles but often face challenges due to computational complexity and uncertainties in process representation. ML models, while effective at capturing complex patterns, can struggle with generalization when trained on limited real-world data. KGML addresses these issues by pretraining ML models on synthetic data generated by PB models, allowing them to learn domain-specific knowledge and fundamental temporal patterns. This pretraining step enhances the model’s ability to recognize key drivers of variability before being fine-tuned with in-situ observational data, which helps correct biases and improve predictive accuracy in real-world applications. KGML has been successfully applied in Earth system and land surface modeling, including hydrological cycle prediction and atmospheric sciences\cite{kraft_towards_2022,he2022improving}. However, despite its success in these domains, no KGML model has yet been developed for estimating wetland CH\textsubscript{4} emissions, leaving a critical gap in leveraging AI-driven approaches for CH\textsubscript{4} flux modeling.
\section{Dataset Construction}

The X-MethaneWet dataset integrates both simulated and observed CH\textsubscript{4} data from different sources and  different scales. 
In the following,  we describe how we create the simulated data using the TEM-MDM model and the observation data from FLUXNET-CH\textsubscript{4}. 

\subsection{Simulated data: TEM-MDM}
\sloppy
TEM-MDM enhances existing PB models by explicitly accounting for permafrost freeze-thaw dynamics and vegetation carbon dynamics, as well as simulating methane production, oxidation, and transport processes on an hourly basis~\cite{zhuang_methane_2004}. We conduct global CH\textsubscript{4} flux intensity simulations using TEM-MDM at a daily and 0.5-degree resolution from 1979 to 2018. The drivers include ERA-Interim climate data~\cite{dee2011era}, vegetation type (\texttt{cltveg})~\cite{melillo1993global}, plant function type (\texttt{vegetation\_type\_11})~\cite{matthews1999recently}, topsoil bulk density  (\texttt{topsoil\_bulk\_density}) ~\cite{nachtergaele2010harmonized}, sand, silt, and clay fractions (\texttt{clfaotxt})~\cite{zhuang2003carbon}, pH (\texttt{phh2o})~\cite{carter1999generating}, wetland types~\cite{matthews1987methane}, elevation (\texttt{clelev})~\cite{toutin2002three}, global annual $\text{CO}_2$ concentration (\texttt{kco2}), global annual $\text{CH}_4$ concentration (\texttt{ch4}). 

Specifically, the feature set includes static, yearly, monthly and daily features, all aligned at the daily level. The climate data are at daily resolution and provide variables such as precipitation (\texttt{PREC}), air temperature (\texttt{TAIR}), solar radiance (\texttt{SOLR}), and vapor pressure (\texttt{VAPR})~\cite{dee2011era}. The intermittent variable net primary productivity (\texttt{NPP}) was generated from TEM-MDM and is on a monthly scale. We also use yearly features such as global annual $\text{CO}_2$ and $\text{CH}_4$ concentration. 
Categorical features, including the wetland distribution map and the spatial distribution of characteristics (e.g., vegetation) are static over time. All these features are duplicated and expanded to daily resolution. 
Detailed descriptions of the above features are provided in Appendix ~\ref{App_TEM}. 

% numerical features are provided in Appendix~\ref{App_TEM}. Categorical features, including the wetland distribution map and the spatial distribution of features (e.g., vegetation), are presented in Figure~\ref{fig:vegetation} in Appendix~\ref{App_TEM}.

% The feature set includes static, yearly, monthly and daily features, all aligned at the daily level. Daily features consist of a 365-day sequence for each location and year. Static features (\texttt{cltveg}, \texttt{clfaotxt}, \texttt{clelev}, \texttt{phh2o}, \texttt{topsoil\_bulk\_density}, \texttt{vegetation\_type\_11}, \texttt{wetland\_type}, \texttt{climate\_type}) and yearly features (\texttt{kco2}, \texttt{ch4}), which have a single value per location and year, are expanded to a 365-day sequence. Monthly feature (\texttt{NPP}) is provided with 12 values per year and location, one for each month, and expanded according to the number of days in each month.
% \textcolor{red}{(which variables are not at daily scale or being static? how do we handle them?)}

Our TEM-MDM dataset consists of 62,470 spatial positions, which are continent locations filtered from the 360 $\times$ 720 grid spanning latitudes [-90°,90°] and longitudes [-180°,180°]. Each position contains daily data over a span of 40 years. We cut the original data into  yearly sequences each consisting of 365 days.  
For the input, we use a 15-dimensional feature space, including the previously mentioned variables.  
More formally, after our processing, the  dimensionality and total volume of  input drivers is given by   
$N_\text{input} = N_S \times N_T \times N_L \times N_D$, where the spatial scale \( N_S \), the temporal scale \( N_T \), the sequence length \( N_L \), and the feature dimension \( N_D \) are 62470, 40, 365, and 15, respectively. 
For the output, we have a single target variable for  CH\textsubscript{4} emission.  
Hence, the dimensionality of our target $\text{CH}_4$ flux is  $N_\text{output} = N_S \times N_T \times N_L$. 

\subsection{Observation data: FLUXNET-CH\textsubscript{4}}
% Impact of Fluxnet-CH4
The FLUXNET-CH\textsubscript{4} Version 1.0 provides standardized data on CH\textsubscript{4} fluxes, measured at 81 sites from multiple regional flux networks using eddy covariance (EC) technology ~\cite{delwiche2021fluxnet}. This half-hourly and daily CH\textsubscript{4} fluxes data covers freshwater, coastal, upland, natural, and managed ecosystems~\cite{delwiche2021fluxnet}. It has been widely used for independent ground truth validation of satellite measurements and Earth system models and improving our understanding of ecosystem-scale CH\textsubscript{4} flux dynamics ~\cite{pastorello2020fluxnet2015, mcnicol_upscaling_2023}.

%Site-level CH\textsubscript{4} flux data were obtained from FLUXNET-CH\textsubscript{4} and follow the CC-BY-4.0 policy~\cite{delwiche2021fluxnet}. 

We select 30 seasonal or permanent vegetated wetland sites covering bog (5), fen (7), marsh (6), swamp (2), salt marsh (3), drained (1) and wet tundra (6) wetland classes and distributed across Arctic-boreal (13), temperate (13), and (sub)tropical (4) climate zones. As shown in Figure~\ref{SiteInfo}, the sites used in this paper include BW-Gum~\cite{helfter_fluxnet-ch4_2020}, US-DPW~\cite{hinkle_fluxnet-ch4_2016},  US-LA1~\cite{krauss_fluxnet-ch4_2019}, US-LA2~\cite{krauss_fluxnet-ch4_2019-1}, US-NC4~\cite{noormets_fluxnet-ch4_2020}, US-Myb~\cite{matthes_fluxnet-ch4_2016}, US-Srr~\cite{bergamaschi_fluxnet-ch4_2018}, US-ORv~\cite{bohrer_fluxnet-ch4_2016}, US-MRM~\cite{schafer2020fluxnet}, US-OWC~\cite{bohrer_fluxnet-ch4_2018}, US-WPT~\cite{chen_fluxnet-ch4_2016}, JP-BBY~\cite{ueyama_fluxnet-ch4_2020}, US-Los~\cite{desai_fluxnet-ch4_2016-1}, FR-LGt~\cite{jacotot_fluxnet-ch4_2020}, DE-SfN~\cite{schmid_fluxnet-ch4_2020}, DE-Zrk~\cite{sachs_fluxnet-ch4_2020}, DE-Hte~\cite{koebsch_fluxnet-ch4_2020}, CA-SCB~\cite{sonnentag_fluxnet-ch4_2019}, FI-Sii~\cite{vesala_fluxnet-ch4_2020}, FI-Si2~\cite{vesala_fluxnet-ch4_2020}, US-BZB~\cite{euskirchen_fluxnet-ch4_2020}, US-BZF~\cite{euskirchen_fluxnet-ch4_2020}, FI-Lom~\cite{lohila_fluxnet-ch4_2020}, US-Ivo~\cite{zona_fluxnet-ch4_2016}, US-ICs~\cite{euskirchen_fluxnet-ch4_2020}, RU-Che~\cite{goeckede_fluxnet-ch4_2020}, RU-Ch2~\cite{goeckede_fluxnet-ch4_2020}, US-Atq~\cite{zona_fluxnet-ch4_2020-2}, US-Bes~\cite{zona_fluxnet-ch4_2020}, and US-Beo~\cite{zona_fluxnet-ch4_2020-1}. Figure~\ref{FLUXNET_time_series} shows an example of $\text{CH}_4$ flux variation in the FLUXNET-$\text{CH}_4$ dataset alongside the two most important features, air temperature and precipitation, illustrating their relationships.

For FLUXNET-$\text{CH}_4$ data, the time span varies across different sites but generally %falls within the year range of  
ranges from 2006 to 2018. The specific time period for each site is shown in Figure~\ref{FLUXNET_time_span} in Appendix \ref{App_fluxnet}. In total, the dataset contains 109 site-years of daily sequence data. The dataset included latitude, longitude, elevation, International Geosphere-Biosphere Programme (IGBP) land cover classification for every site, and temporal information including air temperature, precipitation, vapor pressure, and $\text{CH}_4$ emission labels. For the X-MethaneWet dataset, we select sites classified as `wet' in the IGBP classification column, resulting in a total of up to 30 sites. We use the available values in FLUXNET-$\text{CH}_4$ for data constructions and derive the wetland type from the WAD2M wetland distribution map \cite{zhang2021development}, as shown in Figure~\ref{SiteInfo} in Appendix \ref{App_fluxnet}. For missing features, we fill in the gaps using values from the nearest grid point in TEM-MDM, ensuring that the data is padded to a 15-dimensional feature space consistent with the TEM-MDM dataset. FLUXNET-$\text{CH}_4$ emission approximately range from -100 to 900 mg C m$^{-2}$ d$^{-1}$. 
% More details about the dataset, including the wetland distribution map and the spatial distribution of features (e.g., vegetation), are provided in  Appendix.

% \textcolor{red}{To add: (1) time span, (2) dimensions of observed data, (3) scale difference (4) how we create input drivers for observed dataset}

% \subsection{ML-ready data formats}
% \subsection{ML-ready data structure}

% \textcolor{red}{To add: (1) The dimension of simulated data (input and output); (2) dimensions of observed data, how we create input drivers for observed dataset; (3)  }

\section{Evaluation Framework}

% \subsection{Data usage and ML tasks}
% \textbf{Data usage.}
% We analyzed a total of 64,720 positions worldwide, with the distribution of locations shown in Fig.\ref{}. We use dynamic input features on a daily scale from XXX to XXX, covering a total of XXX days. The input climate features are derived from the TEM dataset\cite{} and include 15 variables related to XXX. The target variable is the methane quantity across all positions.

\subsection{Experimental test cases}
% Problem setting

% regression problem, assume weather covariance available, 

% \textbf{ML tasks.}
An effective CH\textsubscript{4} predictive model should generalize well to unseen scenarios. In practice, the CH\textsubscript{4}  predictive model is often used to make predictions for future periods or in unobserved spatial regions.  
%and generalizability of machine learning models, 
Hence, we perform two different prediction tasks to reflect these real-world settings: temporal extrapolation and spatial extrapolation. The objective of the temporal extrapolation task is to evaluate the ability of a model to predict methane emission for a future period. It is noteworthy that this prediction task is different from typical forecasting %In this prediction task, 
as the input drivers remain available for the testing period. This is to mirror the practical setting of future prediction which relies on input drivers simulated by external models (e.g., climate models). % based on past observations. 
%This setup allows us to analyze how well the model captures temporal dependencies and trends over time. 
For the spatial extrapolation task, the goal is to evaluate the model’s ability to generalize to locations not present in training data. This %ensures that the model is tested on entirely new spatial locations, 
allows us to assess its ability to predict methane emission for completely unobserved spatial regions, e.g., regions that are distant from existing CH\textsubscript{4} flux towers. %across different regions.
Due to the strong spatial heterogeneity, we conduct a multi-fold cross-validation over all the spatial locations for a more comprehensive evaluation. 

\begin{figure*}
    \vspace{-.05in}
    \centering
    \subfigure[Temporal extrapolation: train-test split across years]{
        \includegraphics[width=0.35\textwidth]{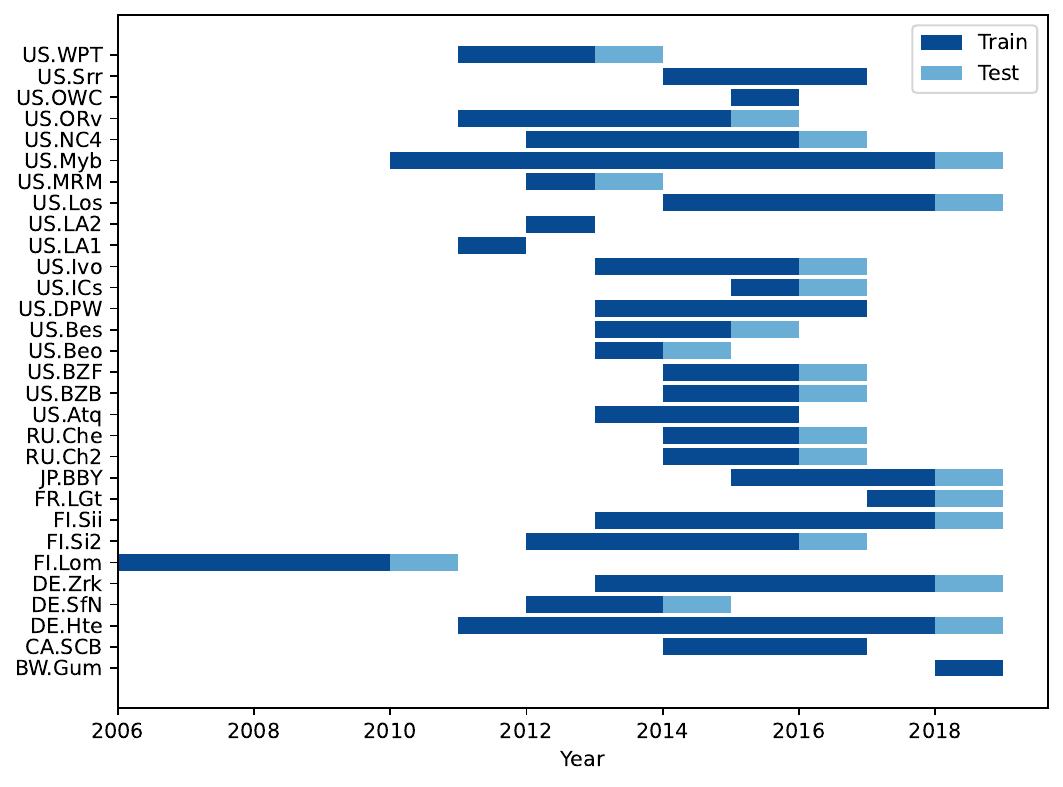}
        \label{FLUXNET_temporal_split}
    }\hspace{.2in}  
    % \hfill
    \subfigure[Spatial extrapolation: train-test split across sites]{
        \includegraphics[width=0.35\textwidth]{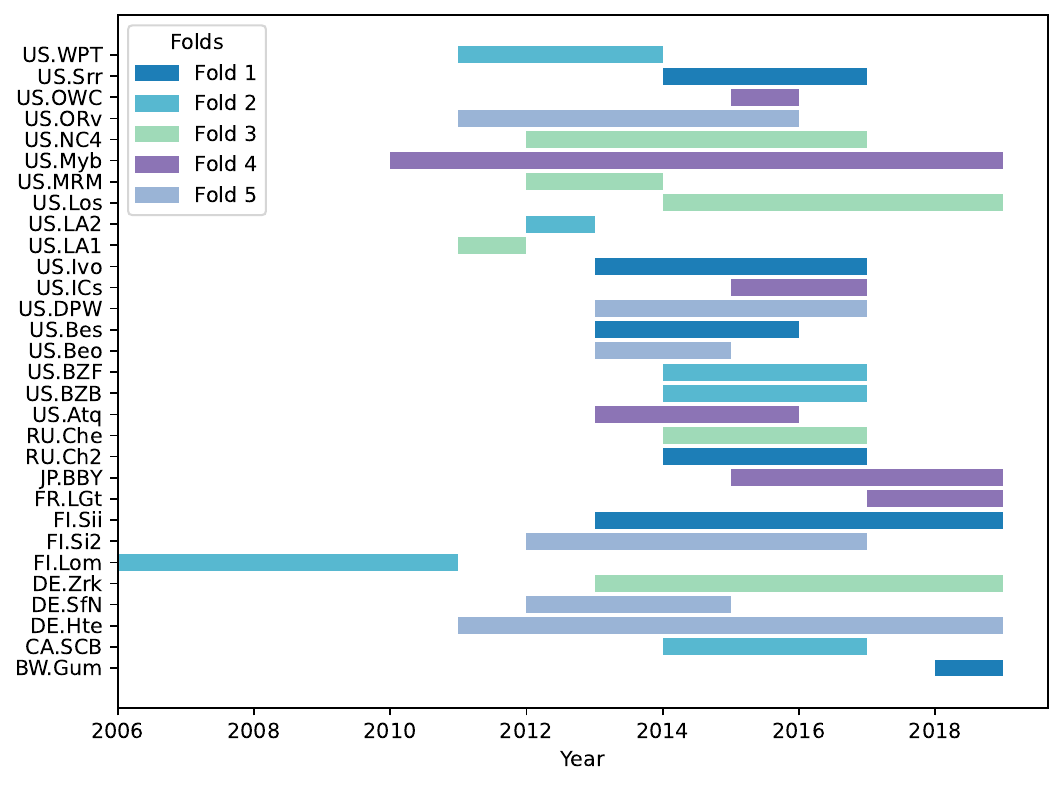}
        \label{FLUXNET_spatial_split}
    }\vspace{-.1in}
    \caption{Train-test split of FLUXNET-$\text{CH}_4$ dataset for temporal and spatial extrapolation experiments.}
    \label{FLUXNET_time_span}
    \vspace{-.05in}
\end{figure*}

Figure~\ref{FLUXNET_time_span} illustrates the train-test splitting strategies applied to the FLUXNET-$\text{CH}_4$ dataset, including temporal extrapolation (split across years) and spatial extrapolation (split across sites), used for evaluating generalization across time and space.

\subsection{Evaluation metrics}

The assessment of predictive performance employs two metrics: normalized root mean squared error (nRMSE) and the coefficient of determination (R$^2$ score). 
Specifically, nRMSE quantifies the average magnitude of prediction errors relative to the mean of the observed data $\bar{y}$ (defined as $\text{nRMSE} = \text{RMSE} / \bar{y}$), allowing for a more meaningful comparison across different datasets.
% Specifically, nRMSE quantifies the average magnitude of prediction errors relative to the scale of the observed data, allowing for a more meaningful comparison across different datasets. % and models. 
A lower nRMSE indicates better predictive accuracy. R$^2$ score evaluates how well the model can use the input drivers to explain the variance in the target CH\textsubscript{4} data. The R$^2$ score has a maximum value of 1, with a higher score indicating a better fit of the model to the data. Between these two metrics, nRMSE accounts for the scale of CH\textsubscript{4} values while R$^2$ considers the variance of CH\textsubscript{4} values. %Both metrics consider the variations in the dataset, which ensures a fair comparison. 

\subsection{Experimental design}

We conduct temporal and spatial extrapolation tasks using both the TEM-MDM-based simulated dataset  and the observed FLUXNET-$\text{CH}_4$ dataset. Specifically, we consider three testing scenarios: % under three different scenarios: 
predicting simulated data, predicting observed data, and transfer learning. For each testing scenario, we will perform both temporal and spatial extrapolation. More details are described as follows.  

% \begin{itemize}[leftmargin=15pt]
    \noindent \textbf{Predicting simulated data}: Using TEM-MDM, in temporal extrapolation experiments, we use data from 1979 to 1998, covering a total of 20 years for training across all locations (a total of 62,470 positions), and evaluate the models on the following period from 1999 to 2018, spanning another 20 years across all locations. For spatial extrapolation tasks, we divide all locations into five equal folds, each containing 12,494 positions. For each run, the models are trained on data from four folds, which include data from 49,976 locations over all 40 years (i.e., 1979 to 2018), and tested on the remaining fold (12,494 positions over 40 years). To ensure robustness, we perform cross-validation by rotating the test fold across different runs.
    
    \noindent \textbf{Predicting observed data}: Using FLUXNET-$\text{CH}_4$, in temporal extrapolation experiments, we hold out the last years of certain sites for testing, while using the remaining years from all sites for training, as shown in Figure~\ref{FLUXNET_temporal_split}. %in Appendix. 
    In spatial extrapolation experiments, we divide all locations into five equal folds, each containing six sites, as shown in Figure~\ref{FLUXNET_spatial_split} %in Appendix
    . The models are trained using data from four folds, covering complete time periods, and tested on the remaining fold. Similar to the previous case, we perform cross-validation by rotating the test fold across different runs.  %More details about the train-test split are provided in the Appendix.
    
    \noindent \textbf{Transfer learning}: 
    Additionally, we conduct experiments to transfer knowledge from simulated data to observed data prediction. Real observations of $\text{CH}_4$ (e.g., the FLUXNET-$\text{CH}_4$ dataset) is often limited in both spatial and temporal scales, posing challenges for robust model training and generalization. To enhance performance on the FLUXNET-$\text{CH}_4$ dataset, we explore transfer learning methods to leverage simulated data from the TEM-MDM dataset for predicting true observations of $\text{CH}_4$ in the  FLUXNET-$\text{CH}_4$ dataset.  Specifically, we split the FLUXNET-$\text{CH}_4$ dataset in the same way as in  \textit{predicting observed data} for both temporal and spatial extrapolation tests, and use  the complete TEM-MDM simulated dataset as the source domain in transfer learning. 

\section{Experiments}
In this section, we conduct multiple experiments using  a range of candidate temporal ML models to demonstrate the promise and gap in methane emission prediction.  %to present the initial analysis.

\begin{figure*}
    % \vspace{-0.1in}
    \centering
    \subfigure[Mean $\text{CH}_4$ emission in 2018]{
        \includegraphics[width=0.32\textwidth]{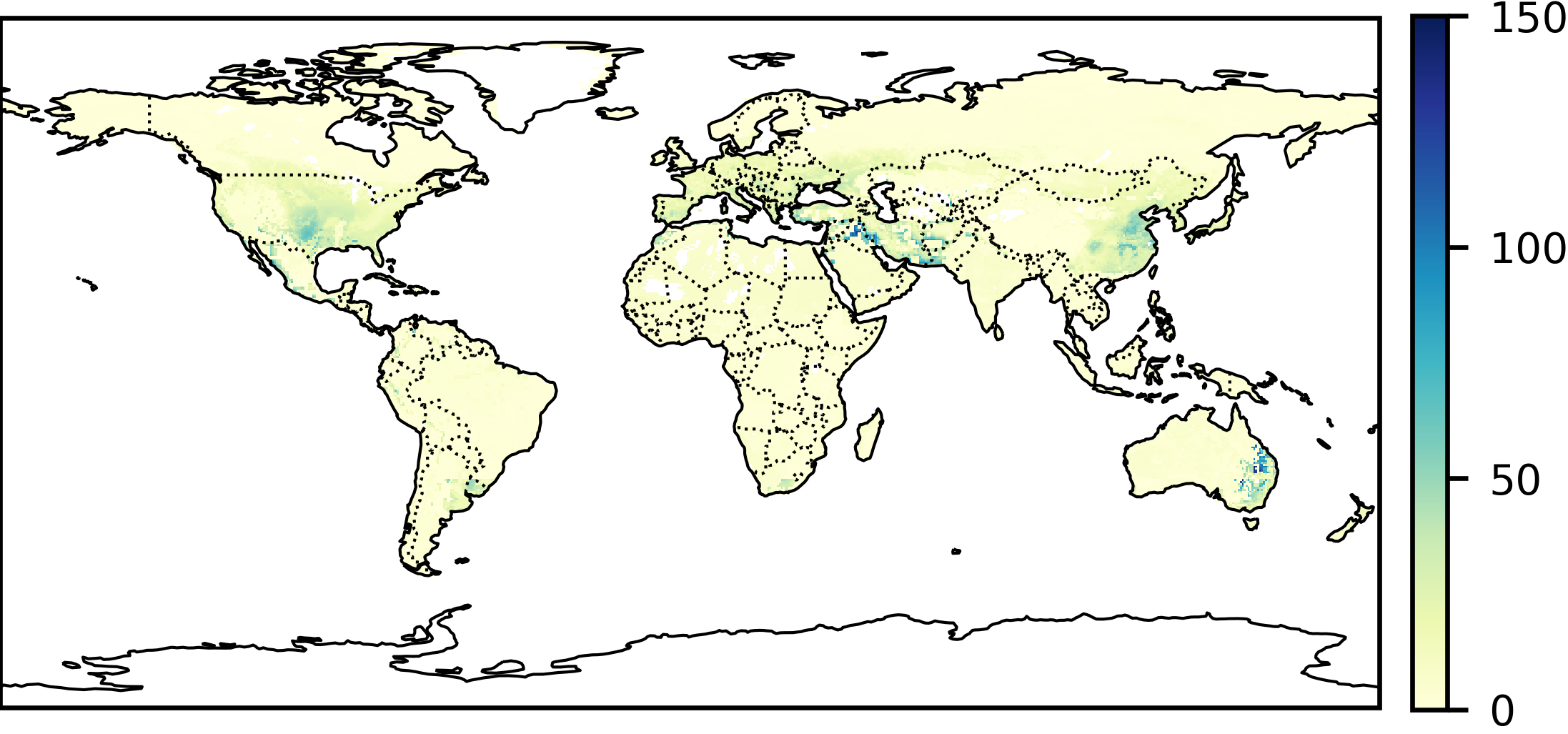}
        \label{fig:TEM_CH4}
    }
    \hfill
    \subfigure[nRMSE of $\text{CH}_4$ prediction for 2018]{
        \includegraphics[width=0.32\textwidth]{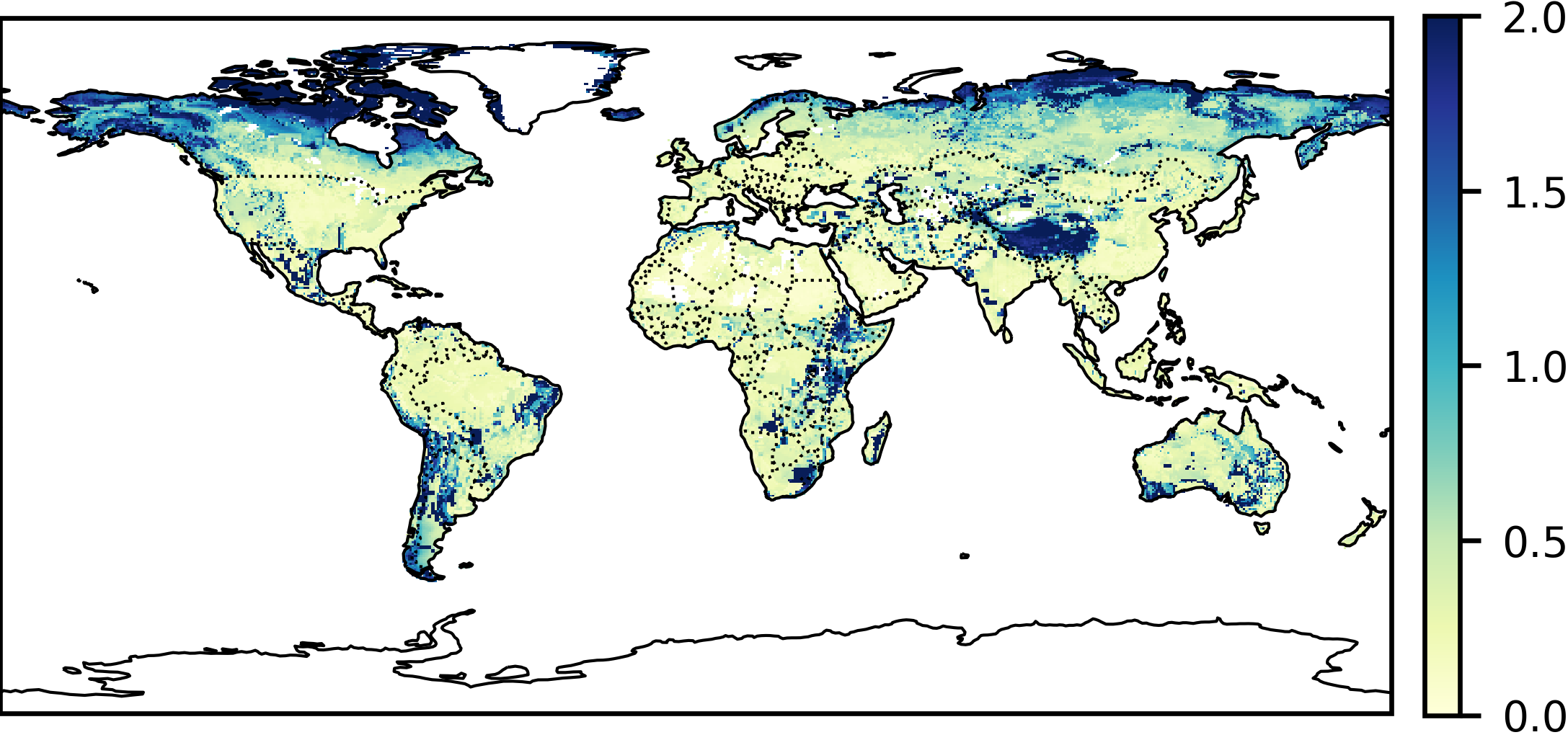}
        \label{fig:TEM_rmse}
    }
    \hfill
    \subfigure[$\text{R}^2$ Score of $\text{CH}_4$ prediction for 2018]{
        \includegraphics[width=0.32\textwidth]{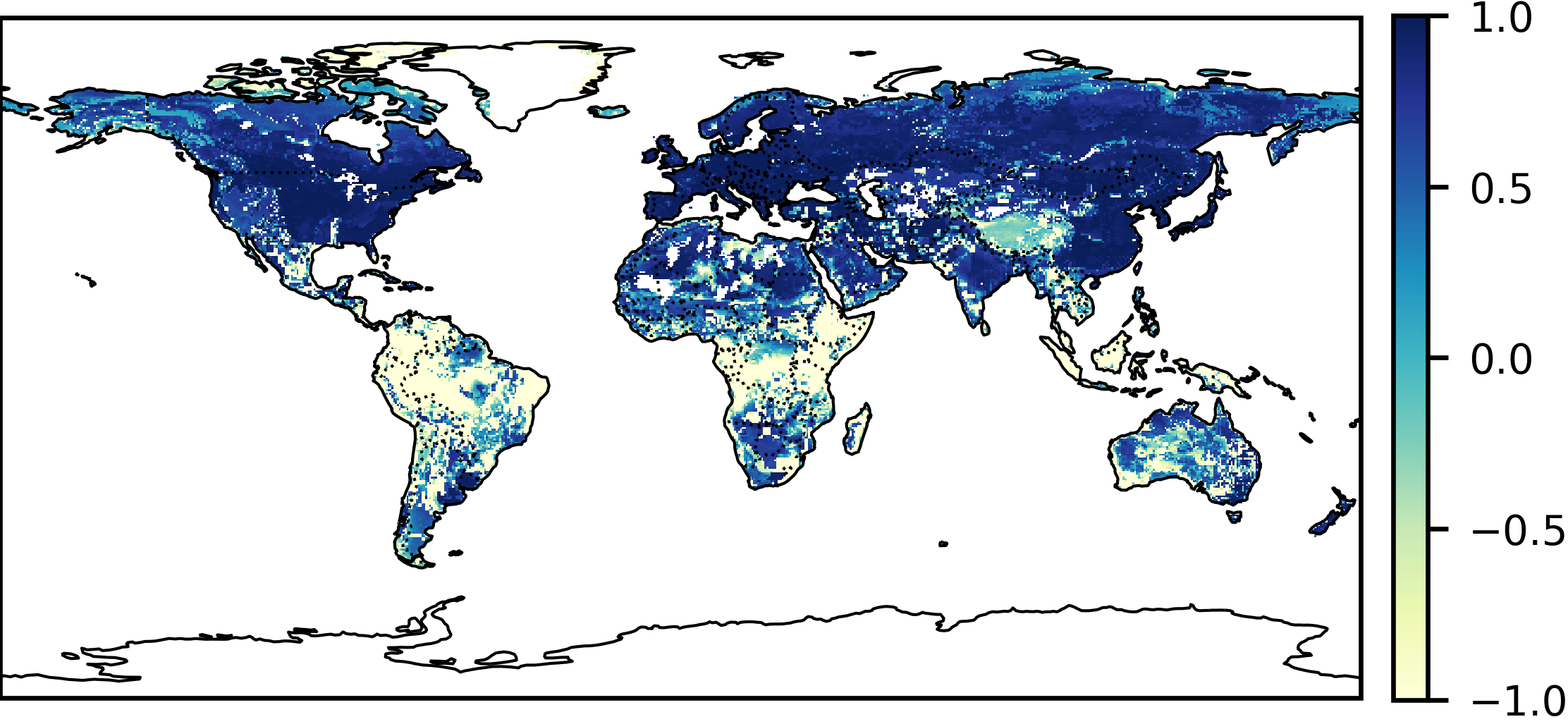}
        \label{fig:TEM_r2}
    }
    \vspace{-0.05in}
    \caption{Label and LSTM prediction performance in the TEM-MDM dataset in 2018.}
    \label{fig:TEM_map}
    \vspace{-0.05in}
\end{figure*}

% \begin{figure*}[htbp]  
%     \centering
%     \includegraphics[width=\textwidth]{figure/global_map_with_timeseries_composite.pdf} 
%     \caption{Illustration of regional heterogeneity in CH$_4$ emission dynamics and predictive performance.The central map shows the spatial distribution of the coefficient of determination (R$^2$) between predicted and observed daily methane emissions, while the surrounding panels illustrate representative regional time series. Each subplot compares observed and predicted emissions for a distinct biogeographical region, highlighting substantial variability in temporal dynamics, emission magnitude, and predictability across regions.}
%     \label{fig:map_ts} 
% \end{figure*}

\begin{figure*}
    % \vspace{-0.15in}
    \centering
    \subfigure[ One example in tropical region: (0.5°, -58.5°)]{
        \includegraphics[height=2.7cm]{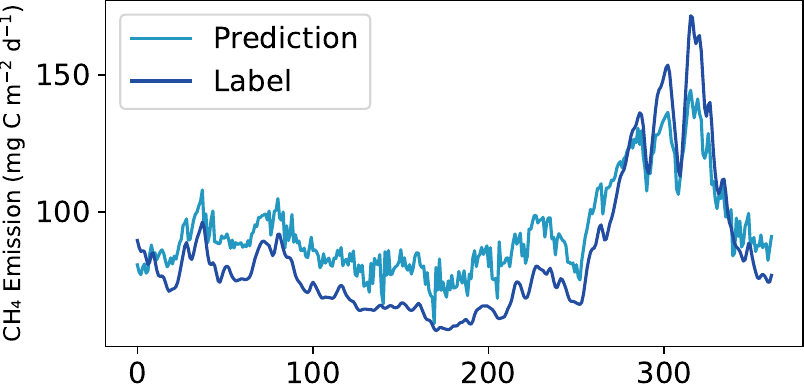}
        \label{fig:TEM_ts_Tropical}
    }
    % \hfill
    \subfigure[ One example in temperate region: (45.0°, 124.0°)]{
        \includegraphics[height=2.7cm]{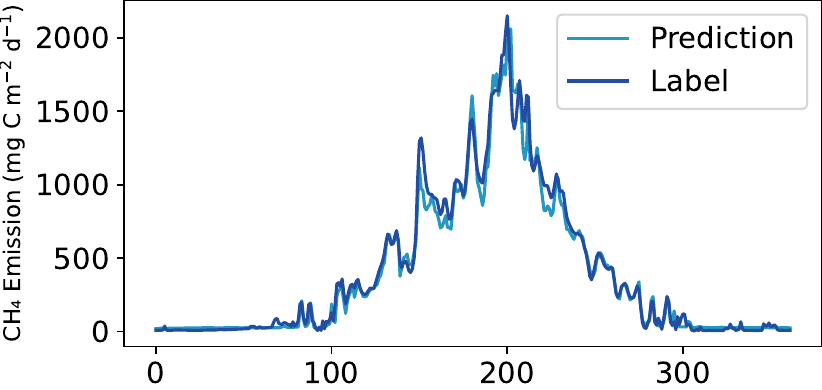}
        \label{fig:TEM_ts_Temperate}
    }
    % \hfill
    \subfigure[ One example in polar region: (77.5°, -112.0°)]{
        \includegraphics[height=2.7cm]{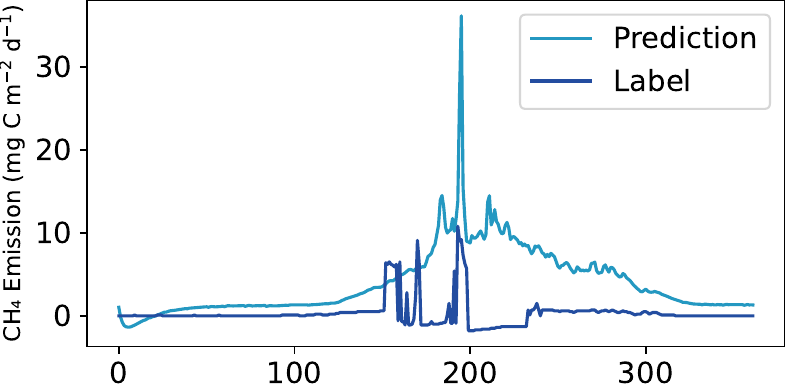}
        \label{fig:TEM_ts_Polar}
    }
    % \vspace{-0.05in}
    \caption{Comparison of LSTM predictions and labels in the TEM-MDM dataset across climate zones in 2018.}
    % \vspace{-0.05in}
\end{figure*}

\subsection{Experimental settings}

% \usepackage{multirow}
% \usepackage{multirow}

\textbf{Candidate methods.} %In this initial analysis, we evaluate 
%the TEM-MDM and FLUXNET-$\text{CH}_4$ datasets using , including 
We test multiple existing temporal ML models, including two traditional LSTM-based methods: LSTM~\cite{LSTM} and EA-LSTM~\cite{kratzert2019}, a temporal CNN-based model: TCN~\cite{liu2019time}, and three encoder-only Transformer-based models: Transformer~\cite{vaswani2017attention}, iTransformer~\cite{liu2023itransformer}, and Pyraformer~\cite{liu2022pyraformer}. Specifically, EA-LSTM is a variant of LSTM commonly used for sequential data prediction in ecological modeling by generating the gating variables using only the static environmental characteristics (e.g., soil properties). The iTransformer and Pyraformer methods are popular methods for time-series modeling tasks, by focusing on inter-variable dependencies and multi-resolution temporal dependencies, respectively. We do not include results for the random forest model in this study, as our results indicate lower performance compared to temporal ML models in predicting daily methane emissions.

%We also test multiple transfer learning methods, %including \textcolor{red}{XXXX}

To effectively transfer information from the source domain (TEM-MDM) to the target domain (FLUXNET-$\text{CH}_4$), we also employ two categories of transfer learning approaches: parameter transfer and data transfer. 
In the parameter transfer category, models are first trained on the source domain and then utilized in the target domain. In our experiments, we implement two methods: the pre-train and fine-tune method, and the residual learning method. In the pre-train and fine-tune method, we first train the model on the TEM-MDM until convergence. We then further fine-tune the pre-trained model using the FLUXNET-$\text{CH}_4$. This approach allows the model to capture the inherent relationships between features and $\text{CH}_4$ emission in the TEM-MDM data, enabling a faster and more effective adaption to the FLUXNET-$\text{CH}_4$ data. Instead of directly fine-tuning, the residual learning method leverages the pre-trained model from TEM-MDM as a baseline for the FLUXNET-$\text{CH}_4$. A separate model is then trained to predict the residual difference between the predicted $\text{CH}_4$ emission from the pre-trained model and the ground-truth $\text{CH}_4$ emission from FLUXNET-$\text{CH}_4$ data. This approach explicitly considers the discrepancies of label spaces, allowing the model to adjust the predictions based on domain-specific variations.

\begin{table}[!b]
\vspace{.05in}
\small
\newcommand{\tabincell}[2]{\begin{tabular}{@{}#1@{}}#2\end{tabular}}
\centering
\caption{Quantitative performance, measured by (nRMSE, R$^2$ score), is evaluated on the TEM-MDM dataset for temporal and spatial extrapolation tasks. The results are averaged over three independent runs. The results in bold highlight the best performance across models.  }
\vspace{0.05in}
\begin{tabular}{|p{1.5cm}|c|c|}
\hline
\textbf{Method} & Temporal & Spatial \\ \hline 
LSTM& (\textbf{0.68$\pm$0.03, 0.92$\pm$0.01})&(\textbf{0.66$\pm$0.04, 0.92$\pm$0.01})\\ 
EALSTM&(0.97$\pm$0.01, 0.84$\pm$0.01)&(0.91$\pm$0.04, 0.86$\pm$0.01)\\ 
TCN&(1.19$\pm$0.04, 0.77$\pm$0.02)&(1.14$\pm$0.07, 0.78$\pm$0.03)\\ 
Transformer&(1.04$\pm$0.03, 0.83$\pm$0.01)&(0.78$\pm$0.03, 0.90$\pm$0.01)\\ 
iTransformer&(1.28$\pm$0.02, 0.74$\pm$0.01)&(1.29$\pm$0.03, 0.73$\pm$0.02)\\  
Pyraformer& (1.05$\pm$0.02, 0.82$\pm$0.01)&(1.13$\pm$0.03, 0.80$\pm$0.01)\\ 
\hline
\end{tabular}
\label{tab:TEM_pretrain}
%\vspace{-0.2in}
\end{table}

In the data transfer category, data from TEM-MDM is directly integrated with FLUXNET-$\text{CH}_4$ data to build a unified model. We implement two methods within this approach: adversarial learning and re-weighting. For adversarial learning, we adopt a method similar to Domain-Adversarial Neural Networks (DANN)~\cite{ajakan2014domain}, where a domain discriminator is trained alongside the main model to minimize the discrepancy between the TEM-MDM and FLUXNET-$\text{CH}_4$ data distributions. The adversarial objective encourages the model to learn domain-invariant features, improving its generalization ability across datasets. Instead of aligning feature distributions through an adversarial objective, the re-weighting method assigns different weights to samples in TEM-MDM to better match the label distribution of FLUXNET-$\text{CH}_4$. The re-weighting is based on statistical properties, specifically the mean and standard deviation of the target data. We compute the differences in mean and standard deviation between TEM-MDM samples and FLUXNET-$\text{CH}_4$ data as a measure of similarity and use these similarity values to re-weight the loss function for each TEM-MDM sample. By dynamically adjusting these weights based on their relevance to the target domain, the model prioritizes FLUXNET-$\text{CH}_4$ data while still leveraging useful information from TEM-MDM. %\textcolor{red}{(add reweighting details.)}

\noindent\textbf{Implementation details.}
We apply data standardization to both the training and testing datasets. Standardization is performed on both the TEM-MDM and FLUXNET-$\text{CH}_4$ datasets using the mean and standard deviation of the TEM-MDM dataset. All candidate methods are implemented using PyTorch 2.5 and run on a GTX 3080 GPU, using the ADAM optimizer, with an initial learning rate of 0.001. All models are set with 3 layers and hidden state dimensions of 8. Transformer-based models use 4 attention heads.

\begin{table}[!t]
%\vspace{-.05in}
\small
\newcommand{\tabincell}[2]{\begin{tabular}{@{}#1@{}}#2\end{tabular}}
\centering
\caption{Quantitative performance, measured by (nRMSE, R$^2$ score), is evaluated on the FLUXNET-$\text{CH}_4$ dataset for temporal and spatial extrapolation tasks. The results are averaged over three independent runs.}
\vspace{0.05in}
\begin{tabular}{|p{1.5cm}|c|c|}
\hline
\textbf{Method} & Temporal & Spatial \\ \hline 
LSTM& (1.02$\pm$0.03, 0.55$\pm$0.03)&(1.40$\pm$0.27, 0.09$\pm$0.17)\\ 
EA-LSTM&(1.14$\pm$0.03, 0.43$\pm$0.01)&(1.45$\pm$0.37, 0.04$\pm$0.31)\\ 
TCN&(1.24$\pm$0.03, 0.32$\pm$0.02)&(1.33$\pm$0.38, 0.18$\pm$0.29)\\ 
Transformer&(1.50$\pm$0.03, 0.03$\pm$0.05)&(1.50$\pm$0.18, -0.01$\pm$0.03)\\ 
iTransformer&(1.44$\pm$0.04, 0.07$\pm$0.02)&(1.38$\pm$0.26, 0.14$\pm$0.18)\\  
Pyraformer& (\textbf{1.02$\pm$0.03, 0.55$\pm$0.01})&(\textbf{1.33$\pm$0.38, 0.19$\pm$0.27})\\ 
\hline
\end{tabular}
\label{table2}
\vspace{.1in}
%\vspace{-0.2in}
\end{table}

\begin{figure*}[htbp]  
    % \vspace{-0.15in}
    \centering
    \subfigure[Fine-tuned vs. scratch-trained model predictions in temporal extrapolation.]{
        \includegraphics[width=0.49\textwidth]{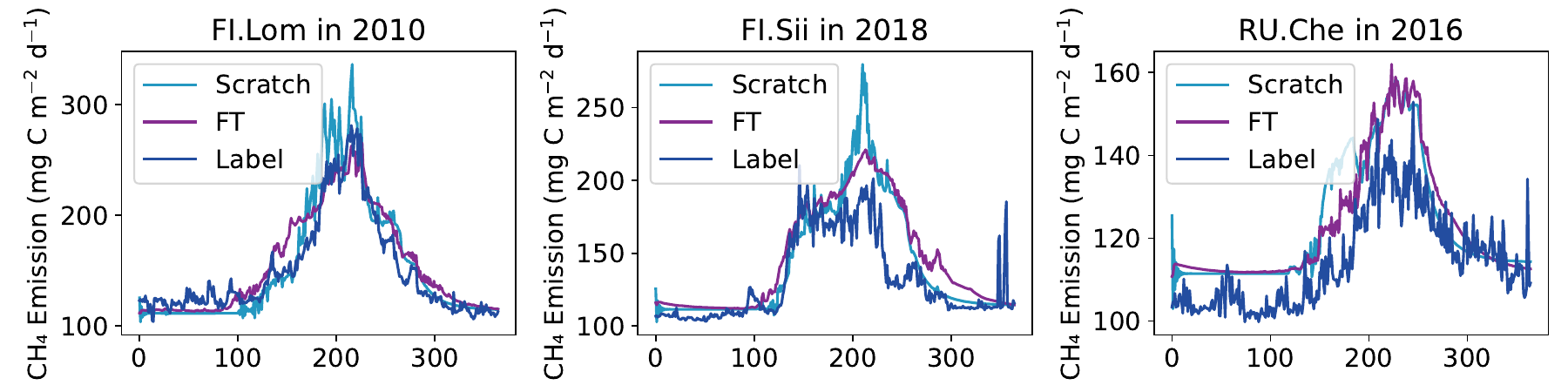}
        \label{fig:FLUXNET_temp_ts}
    }\hspace{-0.1in}
    \subfigure[Fine-tuned vs. scratch-trained model predictions in spatial extrapolation.]{
        \includegraphics[width=0.49\textwidth]{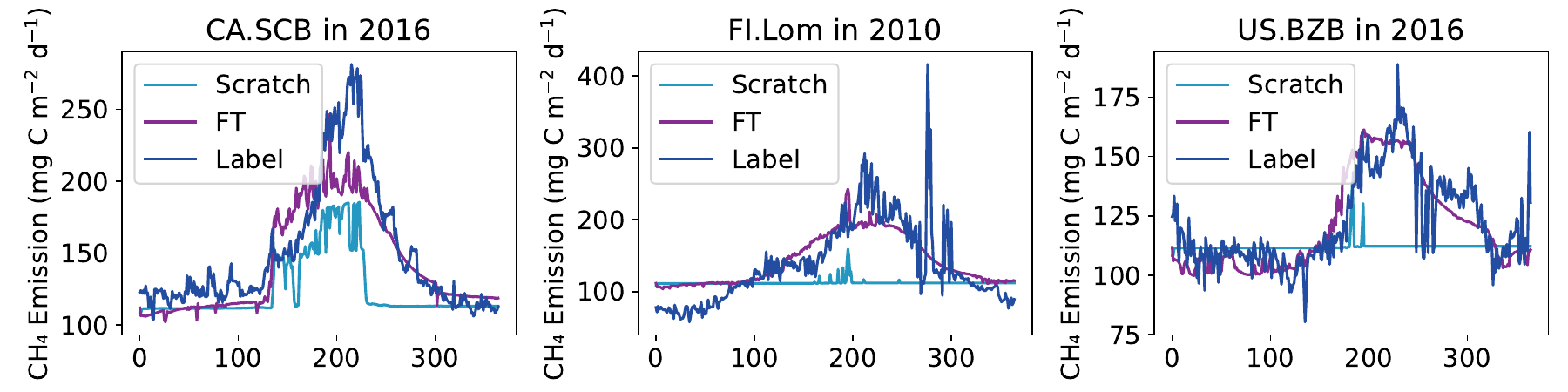}
        \label{fig:FLUXNET_spatial_ts}
    }
    \vspace{-0.05in}
    \caption{Examples showing performance improvements through LSTM fine-tuning in both temporal and spatial extrapolation. } %tasks. } % on the FLUXNET-$\text{CH}_4$ data.}
    \label{fig:FLUXNET_time_series_transfer}
    \vspace{-0.05in}
\end{figure*}

\subsection{Predicting simulated data}
\textbf{Quantitative analysis.} We report the average performance of three independent runs for each candidate method using the simulated TEM-MDM dataset from locations across the globe, as shown in Table~\ref{tab:TEM_pretrain}. Several key observations emerge from the results: (1)~Between the LSTM-based models, LSTM outperforms EA-LSTM in both temporal and spatial extrapolation tasks, achieving lower nRMSE and higher R$^2$ scores. This suggests that the simplified gating mechanism adopted in EA-LSTM  
may not fully capture the factors responsible for modulating the influence of input drivers on wetland states. %do not necessarily lead to better generalization in the simulated data. 
(2) TCN performs poorly in terms of both nRMSE and R$^2$ scores across both tasks. %, indicating that it struggles to capture complex temporal dependencies in methane emissions. 
This suggests that convolutional models are less suitable for sequential methane prediction, likely due to their inherent limitations in modeling long-range dependencies in methane flux dynamics and the lack of adaptive memory mechanisms for modeling complex dynamics. (3) Transformer-based models show varied performance. The standard Transformer model performs worse than LSTM but better than TCN. However, iTransformer and Pyraformer, which introduce specialized architectural improvements, exhibit distinct behaviors. iTransformer has the highest nRMSE and the lowest R$^2$, indicating poor generalization performance. This may be due to its self-attention mechanism operating only across variables rather than across time, which limits its ability to capture methane emission dynamics.   %for temporal extrapolation, indicating poor generalization over time. %However, it performs relatively better in spatial extrapolation (nRMSE = 1.29), suggesting that it may be more effective at capturing spatial variations than temporal trends. 
Pyraformer, on the other hand, demonstrates a more balanced performance, with an nRMSE of 1.05 in temporal extrapolation and 1.13 in spatial extrapolation, placing it between Transformer and iTransformer.

\textbf{Qualitative analysis.} 
To further analyze model performance across different positions within a single year, we evaluate an LSTM model trained on data from 1979 to 2017 and tested on 2018. Figure \ref{fig:TEM_CH4} presents the mean $\text{CH}_4$ emissions in 2018, while Figure~\ref{fig:TEM_rmse} and Figure~\ref{fig:TEM_r2} show the nRMSE and R$^2$ score of LSTM-based $\text{CH}_4$ predictions for the same year. 
From the nRMSE and R$^2$ maps, we observe strong model performance in the temperate region, where both metrics indicate high accuracy. This is further illustrated in Figure~\ref{fig:TEM_ts_Temperate}, where the predictions closely align with the ground truth labels. 
In the tropical region, while nRMSE remains low, the R$^2$ score is poor. As shown in Figure~\ref{fig:TEM_ts_Tropical}, the model's predictions tend to stay close to the target values but exhibit significant oscillations, leading to a low R$^2$ score.
For the polar region, both nRMSE and R$^2$ indicate poor performance. Figure~\ref{fig:TEM_ts_Polar} shows that the predictions deviate noticeably from the target values, highlighting the model’s limitation in this region.

% \textcolor{blue}{
% Beyond these zonal examples, the global maps in Figure~\ref{fig:TEM_map} reveal pronounced regional heterogeneity in both methane emission dynamics and predictive performance. Mid-latitude regions with strong seasonal forcing generally exhibit higher predictability. In contrast, both tropical and high-latitude regions show substantially larger spatial variability and reduced $R^2$ scores, driven by complex hydrological controls and freeze-thaw dynamics respectively. A more detailed regional analysis, including representative time series across diverse biogeographical zones, is provided in Figure~\ref{fig:map_ts} in the Appendix, which further illustrates how ecohydrological heterogeneity shapes model performance across regions.}

% Overall, these results suggest that traditional LSTM models remain strong baselines for time-series methane flux prediction, particularly in capturing long-term dependencies, while Transformer-based models exhibit potential but require further architectural refinement for improved temporal generalization. CNN-based models like TCN appear to be less effective in this context. 

\subsection{Predicting observed data}
%\textbf{Quantitative analysis.} 
We present the average performance of each candidate method using the observed FLUXNET-$\text{CH}_4$ dataset from 30 sites in Table~\ref{table2}. It is worth noting that the standard deviation in the spatial extrapolation setting is relatively large due to the high 
spatial data variability across different FLUXNET-$\text{CH}_4$ sites. This variability significantly impacts the performance in cross-validation, leading to large fluctuations. Furthermore, since the number of observational sites is limited and spatially sparse, test sites in the spatial extrapolation setting are often far from any training sites. This makes generalization particularly difficult and may partly explain the relatively poor results, rather than indicating model overfitting. In addition, there are some observations from the results: (1) For temporal extrapolation, both LSTM and Pyraformer achieve the best results, outperforming other methods with lower nRMSE and higher R$^2$. (2) The standard Transformer and iTransformer struggle in the temporal extrapolation. This suggests the limitations of these models in capturing complex methane dynamics with limited methane flux data, which could involve both long-term and short-term temporal dependencies in practice. (3)~TCN and Pyraformer both show competitive performance for spatial extrapolation.  
However, due to the strong spatial variability (i.e., large standard deviations), they are not consistently better than other models. 
(4) Compared with LSTM, EA-LSTM performs slightly worse, suggesting that its simplified gating mechanism does not necessarily enhance generalization.  
Overall, these observations suggest that Pyraformer remains a strong baseline for methane flux forecasting. 

\begin{table*}[htbp]
\centering
% \caption{Detailed Performance Evaluation: Top row is nRMSE, bottom row is R$^2$ (Mean $\pm$ Std)}
\caption{Detailed quantitative performance evaluation on the FLUXNET-$\text{CH}_4$ dataset. Each cell reports the mean and standard deviation ($\text{mean} \pm \text{std}$) for two metrics: the top value is \textbf{nRMSE} and the bottom value is \textbf{R$^2$}. We highlight the best transfer learning results for each model in the temporal extrapolation setting with \textbf{bold} and the second-best score with \textit{italic}.}
\vspace{0.1in}
\label{tab:final_full_version}
\small
\setlength{\tabcolsep}{3.5pt} 
\begin{tabular}{|l|c|c|c|c|c|c|c|c|c|c|}
\hline
\multirow{3}{*}{Model} & \multicolumn{2}{c|}{\multirow{2}{*}{No Transfer}} & \multicolumn{4}{c|}{Parameter Transfer} & \multicolumn{4}{c|}{Data Transfer} \\ \cline{4-11} 
 & \multicolumn{2}{c|}{} & \multicolumn{2}{c|}{Fine-tune} & \multicolumn{2}{c|}{Residual} & \multicolumn{2}{c|}{Adversarial} & \multicolumn{2}{c|}{Re-weight} \\ \cline{2-11} 
 & temporal & spatial & temporal & spatial & temporal & spatial & temporal & spatial & temporal & spatial \\ \hline

LSTM & 
\makecell{1.02 $\pm$ 0.03 \\ 0.55 $\pm$ 0.03} & \makecell{1.40 $\pm$ 0.27 \\ 0.09 $\pm$ 0.17} & 
\makecell{0.93 $\pm$ 0.07 \\ 0.62 $\pm$ 0.01} & \makecell{1.30 $\pm$ 0.42 \\ 0.20 $\pm$ 0.35} & 
\makecell{0.96 $\pm$ 0.06 \\ 0.59 $\pm$ 0.05} & \makecell{1.33 $\pm$ 0.35 \\ 0.18 $\pm$ 0.22} & 
\makecell{\textit{0.89 $\pm$ 0.02} \\ \textit{0.64 $\pm$ 0.02}} & \makecell{1.38 $\pm$ 0.43 \\ 0.13 $\pm$ 0.35} & 
\makecell{\textbf{0.89 $\pm$ 0.04} \\ \textbf{0.64 $\pm$ 0.03}} & \makecell{1.45 $\pm$ 0.41 \\ 0.03 $\pm$ 0.31} \\ \hline

EA-LSTM & 
\makecell{1.14 $\pm$ 0.03 \\ 0.43 $\pm$ 0.01} & \makecell{1.45 $\pm$ 0.37 \\ 0.04 $\pm$ 0.31} & 
\makecell{0.93 $\pm$ 0.02 \\ 0.60 $\pm$ 0.01} & \makecell{1.35 $\pm$ 0.32 \\ 0.15 $\pm$ 0.22} & 
\makecell{0.96 $\pm$ 0.01 \\ 0.59 $\pm$ 0.01} & \makecell{1.43 $\pm$ 0.29 \\ 0.04 $\pm$ 0.18} & 
\makecell{\textbf{0.80 $\pm$ 0.03} \\ \textbf{0.71 $\pm$ 0.02}} & \makecell{1.50 $\pm$ 0.38 \\ -0.03 $\pm$ 0.31} & 
\makecell{\textit{0.86 $\pm$ 0.02} \\ \textit{0.67 $\pm$ 0.02}} & \makecell{1.60 $\pm$ 0.30 \\ -0.16 $\pm$ 0.59} \\ \hline

TCN & 
\makecell{1.24 $\pm$ 0.03 \\ 0.32 $\pm$ 0.02} & \makecell{1.33 $\pm$ 0.38 \\ 0.18 $\pm$ 0.29} & 
\makecell{1.12 $\pm$ 0.02 \\ 0.43 $\pm$ 0.02} & \makecell{1.40 $\pm$ 0.36 \\ 0.09 $\pm$ 0.30} & 
\makecell{1.15 $\pm$ 0.03 \\ 0.40 $\pm$ 0.03} & \makecell{1.30 $\pm$ 0.29 \\ 0.21 $\pm$ 0.22} & 
\makecell{\textbf{1.05 $\pm$ 0.09} \\ \textbf{0.49 $\pm$ 0.08}} & \makecell{1.50 $\pm$ 0.29 \\ -0.06 $\pm$ 0.25} & 
\makecell{\textit{1.08 $\pm$ 0.03} \\ \textit{0.48 $\pm$ 0.03}} & \makecell{1.58 $\pm$ 0.46 \\ -0.16 $\pm$ 0.47} \\ \hline

Transformer & 
\makecell{1.50 $\pm$ 0.03 \\ 0.03 $\pm$ 0.05} & \makecell{1.50 $\pm$ 0.18 \\ -0.01 $\pm$ 0.03} & 
\makecell{1.15 $\pm$ 0.01 \\ 0.42 $\pm$ 0.01} & \makecell{1.40 $\pm$ 0.42 \\ 0.09 $\pm$ 0.33} & 
\makecell{1.12 $\pm$ 0.03 \\ 0.43 $\pm$ 0.01} & \makecell{1.40 $\pm$ 0.33 \\ 0.10 $\pm$ 0.18} & 
\makecell{\textbf{0.86 $\pm$ 0.01} \\ \textbf{0.66 $\pm$ 0.01}} & \makecell{1.40 $\pm$ 0.29 \\ 0.08 $\pm$ 0.34} & 
\makecell{\textit{0.93 $\pm$ 0.01} \\ \textit{0.61 $\pm$ 0.01}} & \makecell{1.40 $\pm$ 0.36 \\ 0.09 $\pm$ 0.25} \\ \hline

iTransformer & 
\makecell{1.44 $\pm$ 0.04 \\ 0.07 $\pm$ 0.02} & \makecell{1.38 $\pm$ 0.26 \\ 0.14 $\pm$ 0.18} & 
\makecell{\textbf{1.05 $\pm$ 0.03} \\ \textbf{0.50 $\pm$ 0.03}} & \makecell{1.43 $\pm$ 0.39 \\ 0.05 $\pm$ 0.40} & 
\makecell{1.44 $\pm$ 0.07 \\ 0.09 $\pm$ 0.09} & \makecell{1.48 $\pm$ 0.35 \\ -0.01 $\pm$ 0.30} & 
\makecell{1.18 $\pm$ 0.02 \\ 0.38 $\pm$ 0.02} & \makecell{1.38 $\pm$ 0.41 \\ 0.13 $\pm$ 0.33} & 
\makecell{\textit{1.15 $\pm$ 0.06} \\ \textit{0.40 $\pm$ 0.06}} & \makecell{1.43 $\pm$ 0.45 \\ 0.06 $\pm$ 0.44} \\ \hline

Pyraformer & 
\makecell{1.02 $\pm$ 0.03 \\ 0.55 $\pm$ 0.01} & \makecell{1.33 $\pm$ 0.38 \\ 0.19 $\pm$ 0.27} & 
\makecell{\textit{0.86 $\pm$ 0.04} \\ \textit{0.65 $\pm$ 0.03}} & \makecell{1.43 $\pm$ 0.47 \\ 0.05 $\pm$ 0.46} & 
\makecell{0.96 $\pm$ 0.07 \\ 0.59 $\pm$ 0.06} & \makecell{1.28 $\pm$ 0.28 \\ 0.26 $\pm$ 0.17} & 
\makecell{\textbf{0.86 $\pm$ 0.01} \\ \textbf{0.66 $\pm$ 0.01}} & \makecell{1.35 $\pm$ 0.42 \\ 0.15 $\pm$ 0.33} & 
\makecell{0.89 $\pm$ 0.07 \\ 0.64 $\pm$ 0.05} & \makecell{1.35 $\pm$ 0.39 \\ 0.15 $\pm$ 0.29} \\ \hline
\end{tabular}
\end{table*}

\begin{figure}
    % \vspace{-0.05in}
    \centering
    \subfigure[Sparsity test for LSTM]{
        \includegraphics[height=4cm]{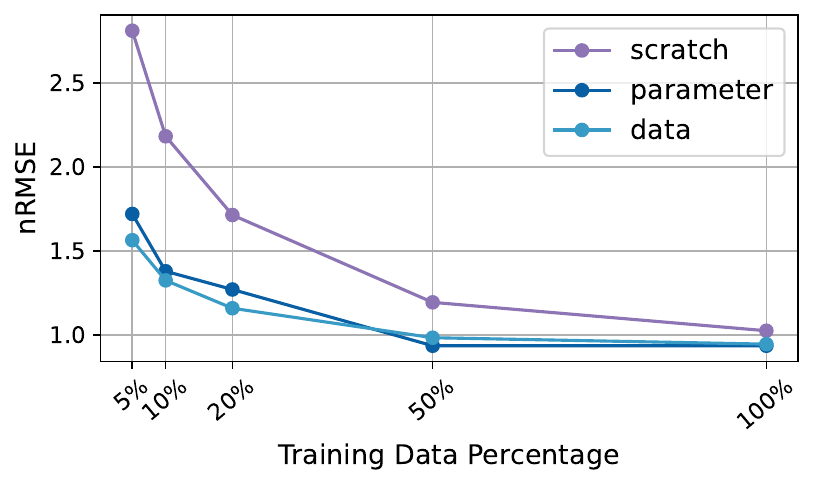}
    }
    \subfigure[Sparsity test for Transformer]{
        \includegraphics[height=4cm]{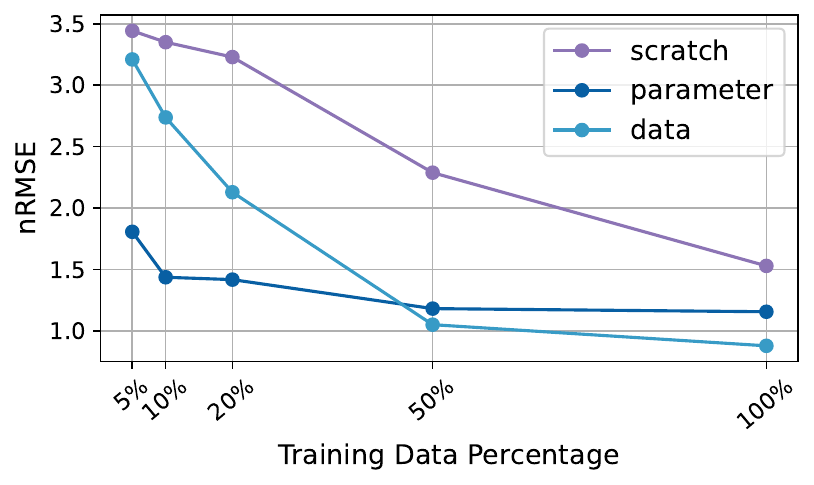}
    }
    \vspace{-0.1in}
    \caption{Sparsity test results in temporal extrapolation task for LSTM and Transformer models. We compare the variants of each model with (1) learning from scratch, (2) data transfer by DANN, and (3) parameter transfer via fine-tuning.  } %three learning scenarios: (a) training from scratch, (b) data transfer using DANN, and (c) parameter transfer via fine-tuning.}
    % \vspace{-0.05in}
    \label{fig:sparse}
\end{figure}

\subsection{Transfer learning from simulations to observations}
\textbf{Quantitative analysis.} 
The experimental results in Table~\ref{tab:final_full_version} highlight the impact of transfer learning approaches on methane flux prediction based on the performance on the FLUXNET-$\text{CH}_4$ dataset. In particular, the ``No Transfer" column represents models trained from scratch using the observed FLUXNET-$\text{CH}_4$ dataset, which is the same as the results reported in Table~\ref{table2}. The remaining columns correspond to various transfer learning strategies. A key observation is that transfer learning generally improves model performance compared to training from scratch, particularly in temporal extrapolation tasks. Most models benefit from transfer learning, showing reduced error and improved predictive accuracy. This suggests that simulated data can help capture fundamental methane flux dynamics, thereby improving model performance on limited observation data. Notably, temporal extrapolation results remain relatively stable across different model initializations, leading to consistently small standard deviations. 
In contrast, spatial extrapolation exhibits substantially higher variability across cross-validation folds. 
This increased variance primarily reflects the strong spatial heterogeneity of methane fluxes: different train–test region splits induce markedly different data distributions, resulting in large performance fluctuations across folds.  
Overall, these results indicate that while temporal dependencies can be learned in a relatively consistent manner, spatial generalization remains considerably more challenging due to pronounced geographical heterogeneity.
% spatial extrapolation exhibits greater instability due to the significant disparity between training and testing data, leading to high variance across different folds in cross-validation. 

Comparing the performance of each model in Table~\ref{tab:final_full_version}, we observe that LSTM performs well even when trained from scratch, demonstrating its effectiveness in handling limited data. LSTM further improves when applying any of the four transfer learning approaches. In contrast, EA-LSTM initially performs worse than LSTM when trained from scratch but significantly improves with transfer learning, especially using data transfer methods. 
A similar pattern is observed in Transformer. Due to its large number of parameters, it is difficult to train effectively with a limited amount of data, resulting in very poor performance when trained from scratch. In contrast, its performance improves significantly with transfer learning from the simulated data, becoming comparable with LSTM-based models. Pyraformer outperforms other Transformer-based models and demonstrates strong performance even when trained from scratch. Such data efficiency is likely attributed to the reduced complexity introduced by the pyramid attention mechanism, which enhances computational efficiency compared with traditional Transformers. Additionally, the hierarchical pyramid structure progressively aggregates information at multiple time scales, enabling Pyraformer to effectively capture long-term dependencies. TCN and iTransformer exhibit the worst performance among all models. Both models fail to adapt well to complex temporal patterns, resulting in significantly lower predictive performance.

Among the transfer learning approaches, fine-tuning achieves the most consistent improvements across both temporal and spatial extrapolation tasks, demonstrating the effectiveness of leveraging simulated data to enhance model generalization. The residual and adversarial training methods show varying degrees of effectiveness, with some models benefiting more than others. These methods appear to improve generalization in certain cases, but their performance is less stable compared to standard fine-tuning. The re-weighting approach, which assigns importance to different training samples, exhibits mixed results, with some models improving while others show a slight degradation in performance, particularly in spatial extrapolation. This indicates that while re-weighting strategies may be beneficial in certain scenarios, they require careful selection of similarity measures to ensure performance improvement. 
Overall, transfer learning proves to be a valuable strategy for methane flux prediction, particularly when models are pre-trained using simulated data before fine-tuning on real observations. These results highlight the importance of combining data-driven approaches and scientific knowledge from physics-based models.

% \textbf{Qualitative analysis.}
% To further analyze the performance of our trained model in both temporal and spatial extrapolation, we visualize several examples for the predictions of the fine-tuned LSTM model 
% on the hold-out years (temporal extrapolation) or sites (temporal extrapolation) 
% While the model successfully captures general patterns, it struggles with finer variations. The difficulty of spatial extrapolation is further amplified by the significant variations in FLUXNET-$\text{CH}_4$ data across different locations. Capturing unseen patterns from entirely new sites proves to be extremely challenging due to the inherent heterogeneity in the space.

\begin{table}[htbp]
\centering
\caption{Relative predictive uncertainty ($std/mean$ ratio) across different models and transfer learning schemes, where lower values indicate higher model confidence.
\textit{None: no transfer; FT: fine-tuning; Res: residual adaptation; Adv: adversarial; RW: re-weighting.}}
\label{tab:uncertainty}
\vspace{0.06in}
\resizebox{\columnwidth}{!}{
\begin{tabular}{|l|ccccc|c|}
\hline
Model & None & FT & Res & Adv & RW & Avg\\ 
\hline
LSTM         & 0.156 & 0.155 & 0.221 & 0.120 & 0.186 & \textbf{0.168} \\
EA-LSTM      & 0.199 & 0.189 & 0.250 & 0.164 & 0.191 & 0.199 \\
TCN          & 0.301 & 0.333 & 0.368 & 0.330 & 0.315 & 0.329 \\
Transformer  & 0.220 & 0.378 & 0.388 & 0.250 & 0.290 & 0.305 \\
iTransformer & 0.302 & 0.243 & 0.336 & 0.147 & 0.128 & 0.231 \\
Pyraformer   & 0.148 & 0.148 & 0.133 & 0.191 & 0.218 & \textbf{0.168} \\
\hline
Average & 0.221 & 0.241 & 0.283 & \textbf{0.200} & 0.221 & \textit{0.233} \\
\hline
\end{tabular}
}
\end{table}
\vspace{0.08in}

\textbf{Qualitative Analysis.} 
To further investigate the predictive behavior of our framework, we visualize representative examples of the fine-tuned LSTM model's predictions for both temporal and spatial extrapolation tasks (Figure~\ref{fig:FLUXNET_time_series_transfer}). 
In the temporal extrapolation task (predicting held-out years at seen sites), the model effectively captures the seasonal phenology and general flux magnitude. However, in the spatial extrapolation task (predicting held-out sites), the performance drops, reflecting the intrinsic difficulty of generalising to unseen geographical contexts. 
This performance gap is primarily driven by the extreme spatial heterogeneity of the FLUXNET-$\text{CH}_4$ dataset, where methane emission patterns vary significantly across different wetland types and microclimates.

\textbf{Uncertainty and reliability analysis.}
While our main experimental results report the ensemble-based uncertainty (mean $\pm$ std) derived from independently trained models, it is equally critical to quantify the intrinsic predictive confidence of each individual model. To this end, we further employ Monte Carlo Dropout ($p=0.2$ with 100 stochastic forward passes) at inference time to compute the relative predictive uncertainty ($std/mean$ ratio). This metric, reported in Table~\ref{tab:uncertainty}, reflects the model's internal stability when facing data perturbations.

The quantitative results reveal significant insights into the reliability of transfer learning. On the strategy level, adversarial learning consistently achieves the lowest average uncertainty across all architectures, suggesting that explicitly aligning the feature distributions of simulations and observations effectively filters out simulation biases, leading to more confident predictions. Regarding model architectures, both Pyraformer and LSTM emerge as the most robust baselines. While LSTM benefits from its architectural simplicity, Pyraformer's robustness highlights the efficacy of its hierarchical pyramid attention in regularizing long-range temporal dependencies. Notably, we observe a `stability-accuracy gap' in models like iTransformer, which can achieve low specific uncertainty ($0.128$ under Re-weighting) but suffer from poor predictive accuracy (Table~\ref{tab:final_full_version}). This indicates that such models may converge to consistently biased minima rather than capturing the high-frequency flux dynamics, underscoring that low uncertainty must be paired with high accuracy to serve as a valid benchmark.

\subsection{Sparsity test in transfer learning}

To mimic real-world challenges in data collection, we also conduct a series of sparsity tests in the temporal extrapolation task. We used 50\%, 20\%, 10\%, and 5\% of the original FLUXNET-$\text{CH}_4$ training data to train both LSTM and Transformer models across three learning scenarios: (1) training from scratch, (2) data transfer using DANN, and (3) parameter transfer via fine-tuning. This approach demonstrates that these models can still exhibit promising generalizability even when the training data is severely limited.

According to the results shown in Figure~\ref{fig:sparse}, we have some observations: (1) Both LSTM and Transformer degrade when less data is available, especially at the 5\% level; (2) both transfer methods (data transfer and parameter transfer) outperform training from scratch, with the most significant gains at 10\% and 5\%; (3) parameter transfer via fine-tuning serves as a strong middle ground, bridging the gap between scratch and data approaches in moderate to low data scenarios; (4) the data transfer approach (DANN) used in LSTM yields the lowest error across all training percentages, even at 5\%;  (5) while the Transformer does well with sufficient data, it is more sensitive to data sparsity than the LSTM, though both architectures benefit greatly from transfer strategies rather than training from scratch. Overall, these findings confirm that both transfer learning approaches (i.e., DANN and transferring pre-trained parameters) effectively reduce the impact of limited observation data in training. This offers new insights into the use of advanced ML models % suggesting that LSTM and Transformer models can remain adaptable 
in real-world situations where large observation data are unavailable. % difficult to obtain.

\section{Conclusion and Future Work}

This paper introduces X-MethaneWet as  the first integrative wetland CH\textsubscript{4} dataset combining both TEM-MDM-based simulations and true observations of methane emissions. The dataset is collected at a daily time interval across the entire globe while maintaining distinct scales between the simulated and observed data. We create a standardized evaluation framework and also conduct extensive experiments using multiple types of ML models. Our results demonstrate the promise of certain ML models in generalizing to unobserved spatial regions and future time periods. The transfer learning experiment also underscores the potential of a hybrid learning paradigm that integrates physical simulations with data-driven modeling of truly observed methane emissions. 

% Limitation
We would like to acknowledge the limitations of this study. Firstly, the global CO\textsubscript{2} and CH\textsubscript{4} concentration and NPP used in this study are at annual or monthly temporal resolution due to the data availability. Also, soil properties and vegetation are assumed static because they are temporally stable and less likely to change over decades unless there is significant land disturbance. In the future, augmenting these variables to a higher temporal resolution could further improve the model's accuracy. Secondly, this study makes use of existing wetland eddy covariance sites, although observational coverage is sparse in tropical regions. We advocate for the establishment of new observation sites in tropical regions, which would significantly improve CH\textsubscript{4} emission estimates~\cite{zhu2024critical}.

% future work
The X-MethaneWet dataset could serve as an important resource to catalyze the exploration of new data mining and machine learning algorithms for predicting global methane emissions. This can greatly facilitate the development of tools for decision making in methane mitigation. 
Future work includes integrating other methane data sources, such as top-down atmospheric inversions~\cite{upton2024constraining}, 
and exploring advanced knowledge-guided learning methods for combining physics-based models and ML algorithms.

\begin{acks}
% This work was supported by National Science Foundation (NSF) grants 2239175, 2316305, 2147195, 2425844, 2425845, 2430978, 2126474, 2530609, 2530610, and 2203581; USGS awards  G21AC10564 and G22AC00266; NASA grants 80NSSC24K1061 and 80NSSC25K0013; and NSF NCAR's Derecho HPC system. This research was also supported in part by the University of Pittsburgh Center for Research Computing through the resources provided. 

This work was supported by the National Science Foundation (NSF) grants 2239175, 2316305, 2147195, 2425844, 2425845, 2430978, 2126474, 2530609, 2530610, and 2203581, and 2153040; USGS awards  G21AC10564 and G22AC00266; NASA grants 80NSSC24K1061 and 80NSSC25K0013; Department of Energy (DOE) grant DE-SC0024360; USDA and NSF funded AI institute AI-LEAF: 2023-67021-39829; and NSF NCAR's Derecho HPC system. This research was also supported in part by the University of Pittsburgh Center for Research Computing.
\end{acks}

\bibliographystyle{ACM-Reference-Format}
\bibliography{sample-sigconf}

\appendix
\begin{figure*}[htbp]
    \centering
    % Adjust each subfigure with \raisebox
    \subfigure[Wetland type]{
        \raisebox{-\height}{\includegraphics[width=0.45\textwidth]{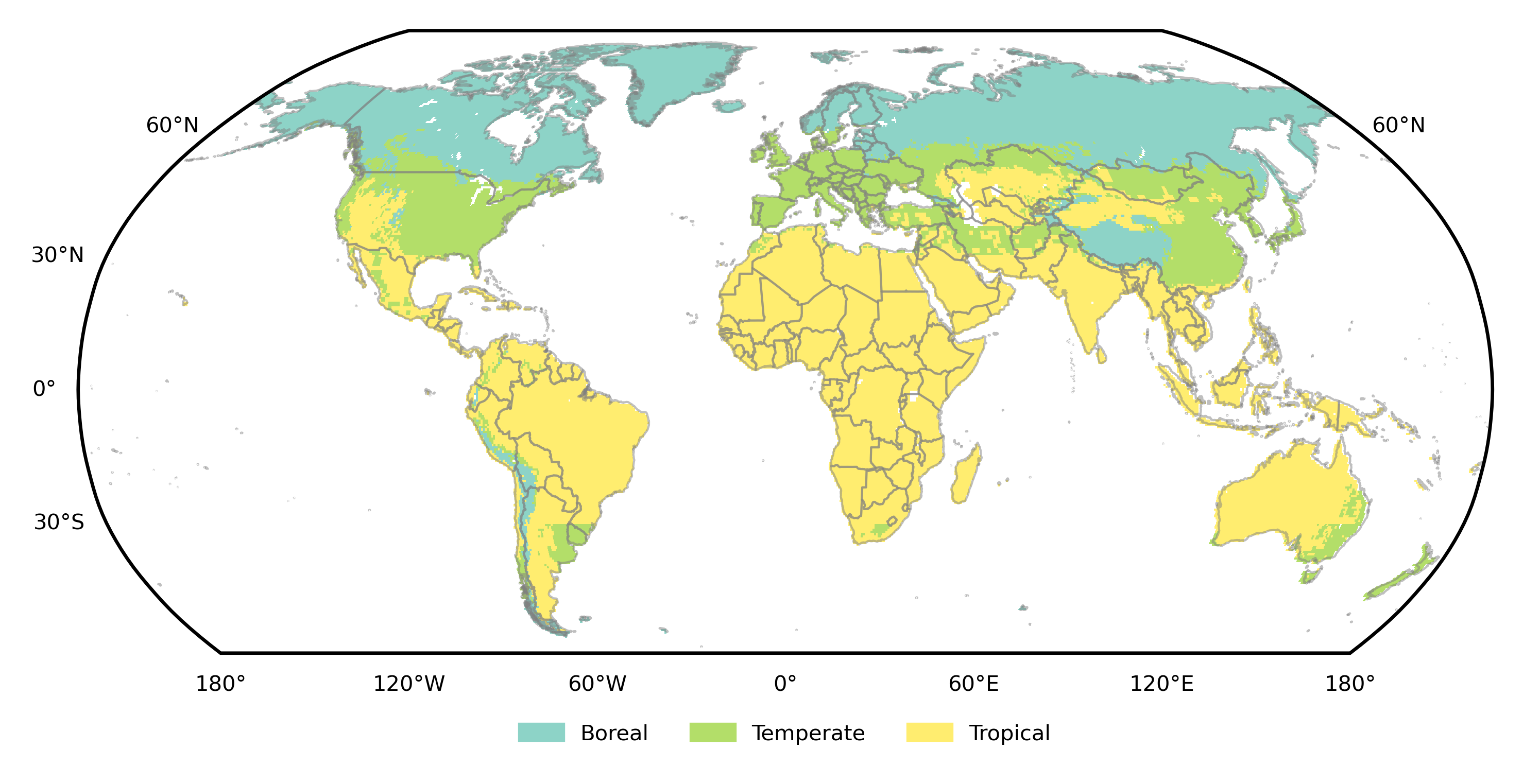}}
    }
    % No need for \hfill if you want them to be close to each other, or adjust spacing as needed
    \subfigure[Climate type]{
        \raisebox{-\height}{\includegraphics[width=0.45\textwidth]{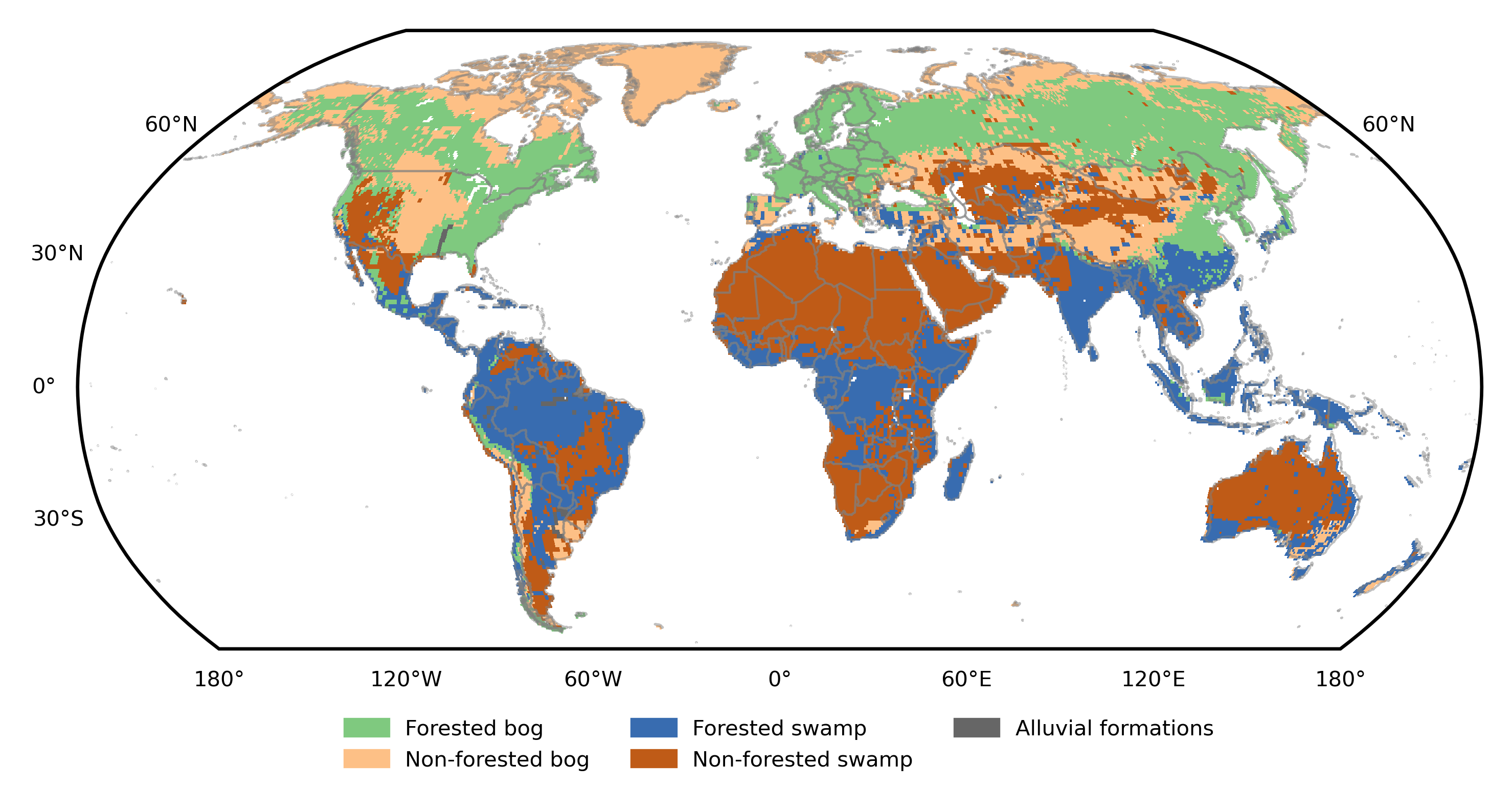}}
    }
    \subfigure[Vegetation function type]{
        \raisebox{-\height}{\includegraphics[width=0.45\textwidth]{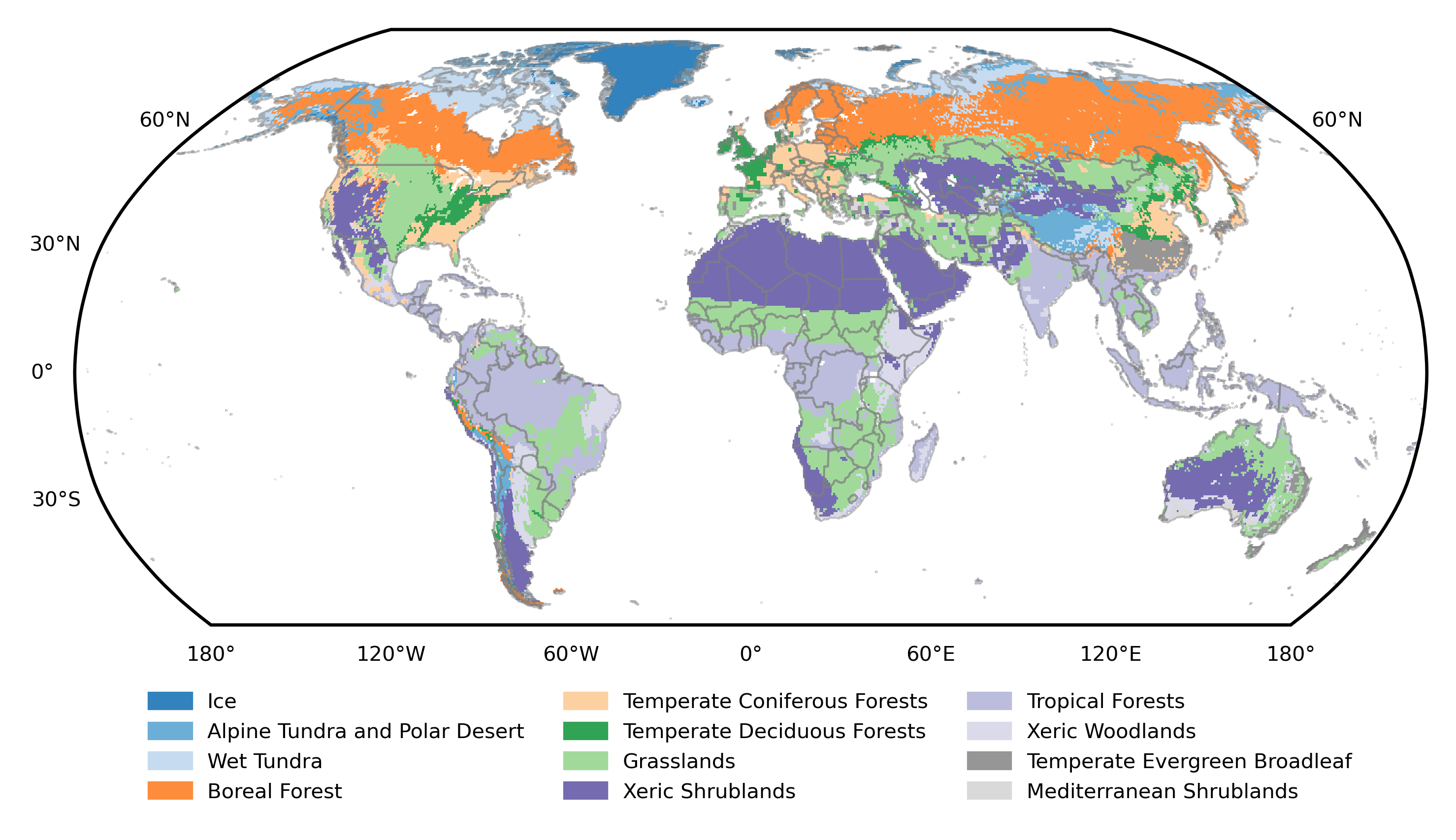}}
    }
    \subfigure[Vegetation type]{
        \raisebox{-\height}{\includegraphics[width=0.45\textwidth]{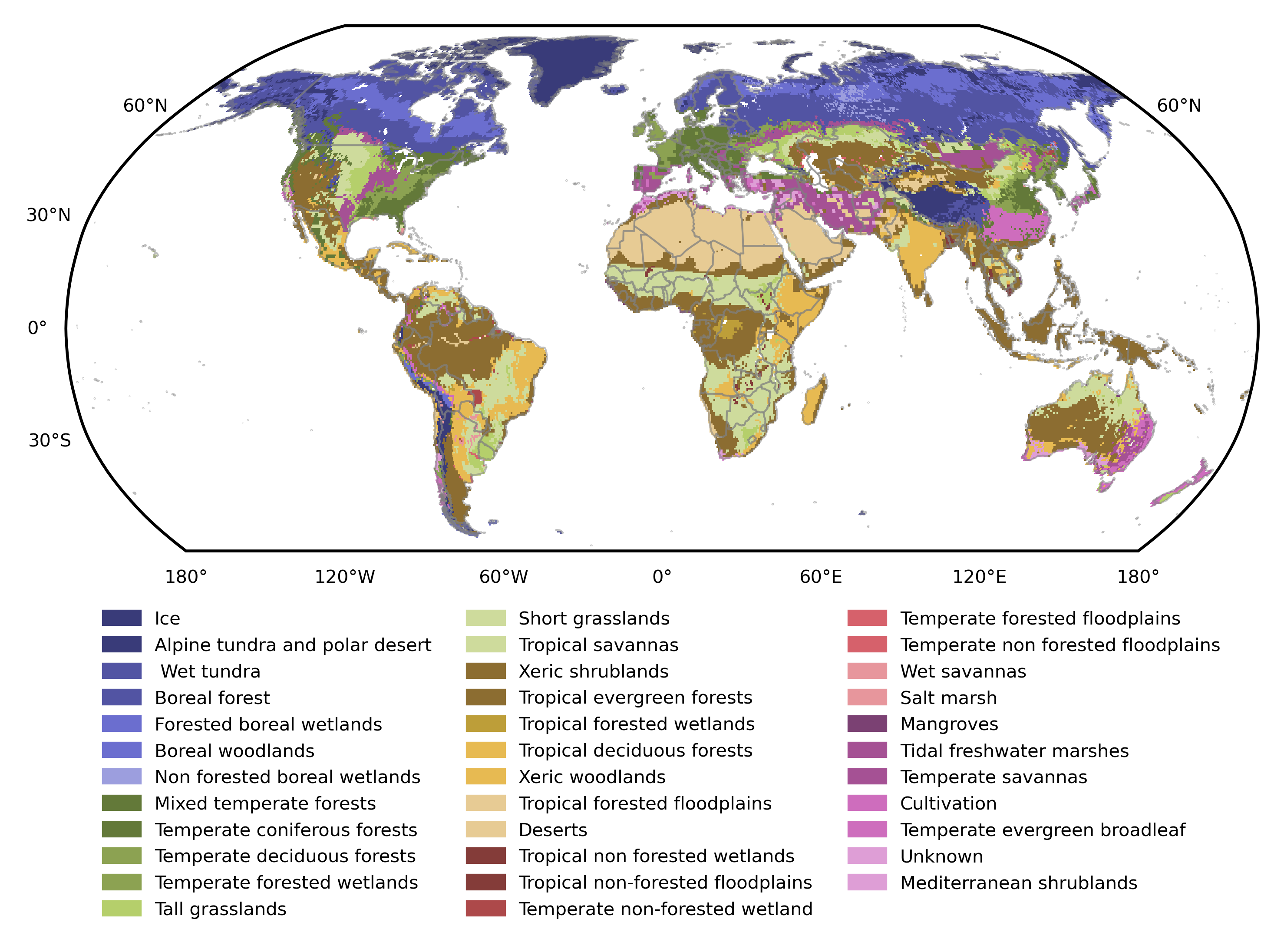}}
    }
    \caption{Spatial distribution of wetland type, climate type, vegetation function type and vegetation type.}
    \label{fig:vegetation}
\end{figure*}

\begin{figure*}[htbp]  
    \centering
    \includegraphics[width=0.9\textwidth]{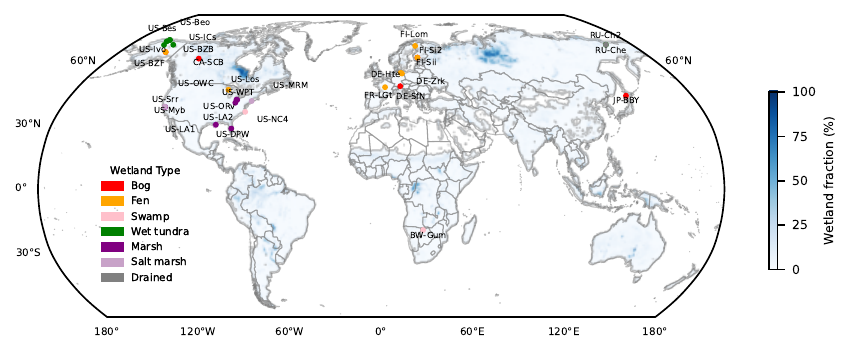} 
    \caption{Location and wetland type of 30 FLUXNET-CH\textsubscript{4} sites. Symbol colors represent the wetland type. The wetland fraction was derived from the WAD2M wetland distribution map~\cite{zhang2021development}.}
    \label{SiteInfo} 
\end{figure*}

% % \begin{figure*}
% %     \centering
% %     \subfigure[Wetland type]{
% %         \includegraphics[height=4cm]{figure/climatetype.pdf}
% %         \label{fig:cltveg}
% %     }
% %     % \hfill
% %     \subfigure[Climate type]{
% %         \includegraphics[height=4.4cm]{figure/wetlandtype.pdf}
% %         \label{fig:TEM_ts_Temperate}
% %     }
% %     \vspace{-0.2in}
% %     % \hfill
% %     \subfigure[vegetation type)]{
% %         \includegraphics[height=4cm]{figure/cltveg.pdf}
% %         \label{fig:TEM_ts_Polar}
% %     }
% %     % \hfill
% %     \subfigure[Vegetation function type]{
% %         \includegraphics[height=4cm]{figure/TEM_ts_Polar.pdf}
% %         \label{fig:TEM_ts_Polar}
% %     }    
% %     \vspace{0in}
% %     \caption{Working on this}
% % \end{figure*}

% \begin{figure*}[htbp]  
%     \centering
%     \includegraphics[width=\textwidth]{figure/climatetype.pdf} 
%     \caption{Spatial distribution of climate type.}
%     \label{climatetype} 
% \end{figure*}

% \begin{figure*}[htbp]  
%     \centering
%     \includegraphics[width=\textwidth]{figure/wetlandtype.pdf} 
%     \caption{Spatial distribution of wetland type.}
%     \label{climatetype} 
% \end{figure*}

% \begin{figure*}[htbp]  
%     \centering
%     \includegraphics[width=\textwidth]{figure/cltveg.pdf} 
%     \caption{Spatial distribution of vegetation type.}
%     \label{climatetype} 
% \end{figure*}

% \begin{figure*}[htbp]  
%     \centering
%     \includegraphics[width=\textwidth]{figure/climatetype.pdf} 
%     \caption{Spatial distribution of wetland type.}
%     \label{climatetype} 
% \end{figure*}

\section{Detailed descriptions of TEM-MDM features}
\label{App_TEM}
To better understand the characteristics of the TEM-MDM dataset, we provide a detailed breakdown of its features:

\begin{table}[htbp]
\centering
% \caption{Summary statistics of input variables used in this study.}
\label{tab:input_stats}
\small
\begin{tabular}{|l|c c c|}
\hline
Variable (Symbol, Unit) & Min & Max & Mean / Median \\
\hline
Precipitation (\texttt{PREC}, mm/day) 
    & 0 & 534 & 2 \\

Air temperature (\texttt{TAIR}, $^\circ$C) 
    & $-71$ & 47 & 9 \\

Solar radiance (\texttt{SOLR}, W/m$^{-2}$) 
    & 0 & 452 & 175 \\

Vapor pressure (\texttt{VAPR}, hPa) 
    & 0 & 46 & 10 \\

Topsoil bulk density (kg/dm$^{3}$) 
    & 0 & 2 & 1 \\

Soil texture fraction (\texttt{clfaotxt}, \%) 
    & 25 & 80 & 49 \\

Soil pH (\texttt{phh2o}) 
    & 0 & 9 & 6 \\

Elevation (\texttt{clelev}, m) 
    & $-51$ & 6130 & 671 (median) \\

\makecell[l]{Net primary productivity (\texttt{NPP},\\ gC m$^{-2}$ month$^{-1}$)}
    & $-387$ & 315 & 25 \\
\hline
\end{tabular}
\end{table}

% \begin{itemize}[leftmargin=15pt]
%     \item Precipitation (\texttt{PREC}) ranges from 0 mm/day to 534 mm/day, with an average of 2 mm/day.
%     \item Air temperature (\texttt{TAIR}) ranges from -71$^\circ$C to 47$^\circ$C, with an average of 9$^\circ$C.
%     \item Solar radiance (\texttt{SOLR}) ranges from 0 W/m$^{-2}$ to 452 W/m$^{-2}$, with an average of 175 W/m$^{-2}$.
%     \item Vapor pressure (\texttt{VAPR}) ranges from 0 hPa to 46 hPa, with an average of 10 hPa.
%     \item Topsoil bulk density ranges from 0 kg/dm$^{-3}$ to 2 kg/dm$^{-3}$, with an average of 1 kg/dm$^{-3}$.
%     \item Sand, silt, and clay fraction (\texttt{clfaotxt}) ranges from 25\% to 80\%, with an average of 49\%.
%     \item PH (\texttt{phh2o}) ranges from 0 to 9, with an average of 6.
%     \item Elevation (\texttt{clelev}) ranges from -51 meters to 6130 meters, with a median of 671 meters.
%     \item Net primary productivity (\texttt{NPP}) ranges from -387 gC m$^{-2}$ month$^{-1}$ to 315 gC m$^{-2}$ month$^{-1}$, with an average of 25 gC m$^{-2}$ month$^{-1}$.
% \end{itemize}

The spatial distribution of categorical vegetation type, plant function type, wetland type, and climate type are shown in Figure~\ref{fig:vegetation}.

\section{Detailed descriptions of 30 FLUXNET-CH\textsubscript{4} sites}
\label{App_fluxnet}

Figure~\ref{SiteInfo} presents the geographic locations and wetland types of 30 FLUXNET-CH$_4$ sites. From the figure, we observe that FLUXNET-CH$_4$ sites span multiple continents and are unevenly distributed worldwide, with some locations being geographically isolated. Additionally, the distribution of wetland types varies across sites, indicating regional deviations in wetland ecosystems.

\end{document}